\documentclass[smallcondensed]{svjour3}

\usepackage[margin={2.5cm,2.5cm}]{geometry}
\usepackage{amsmath}
\usepackage{amsfonts}
\usepackage{amssymb}
\usepackage{mathtools}
\usepackage{bm}
\usepackage{url}
\usepackage{multirow}
\usepackage{latexsym}
\usepackage{booktabs}
\usepackage{algorithm}
\usepackage{algpseudocode}
\usepackage{csvsimple}
\usepackage{siunitx}
\usepackage{pgfplotstable}
\DeclareMathOperator*{\argmin}{argmin}

\title{Interpretable Time Series Classification using All-Subsequence Learning and  Symbolic Representations in Time and Frequency Domains}
\titlerunning{Interpretable Time Series Classification using All-Subsequence Learning}        

\author{Thach Le Nguyen \and Severin Gsponer \and Iulia Ilie \and  Georgiana Ifrim}

\institute{Insight Centre for Data Analytics, University College Dublin, Ireland \\
           \email{\{thach.lenguyen,severin.gsponer,iulia.ilie,georgiana.ifrim\}@insight-centre.org}             
}
\date{Received: date / Accepted: date}

\begin{document}
\maketitle

\begin{abstract}
The time series classification literature has expanded rapidly over the last decade, with many new classification approaches published each year. 
The research focus has mostly been on improving the accuracy and efficiency of classifiers, while their interpretability has been somewhat neglected.
Classifier interpretability has become a critical constraint for many application domains and the introduction of the 'right to explanation' GDPR EU legislation in May 2018 is likely to 
further emphasize the importance of explainable learning algorithms. 
In this work we analyse the state-of-the-art for time series classification, and propose new algorithms that aim to maintain the classifier accuracy and efficiency, but 
keep interpretability as a key design constraint. 
We present new time series classification algorithms that advance the state-of-the-art by implementing the following 
three key ideas: 
(1) \textbf{Multiple resolutions of symbolic approximations:} 
we combine symbolic representations obtained using different parameters, rather than one fixed representation (e.g., multiple SAX representations);  
(2) \textbf{Multiple domain representations:} 
we combine symbolic approximations in time (e.g., SAX) and frequency (e.g., SFA) domains, 
to be more robust across problem domains;  
(3) \textbf{Efficient navigation of a huge symbolic-words space:}
we adapt a symbolic sequence classifier named SEQL, to make it work with multiple domain representations (e.g., SAX-SEQL, SFA-SEQL), 
and use its greedy feature selection strategy to effectively filter the best features for each representation.
We show that a multi-resolution multi-domain linear classifier, SAX-SFA-SEQL, achieves a similar accuracy to the state-of-the-art COTE ensemble, and 
to a recent deep learning method (FCN), but uses a fraction of the time required by either COTE or FCN. 
We discuss the accuracy, efficiency and interpretability of our proposed algorithms.
To further analyse the interpretability aspect of our classifiers, we present a case study on an ecology benchmark.
\end{abstract}

\section{Introduction}
\label{sec:intro}

The introduction of Symbolic Aggregate Approximation (SAX) by Lin et al. \cite{lin-sax:dmkd07} has been the inspiration of numerous studies on time series analysis (e.g.,~\cite{castro:motif,chen:palmprint,kasten:acoustic,costa:patterns}). SAX is a symbolic representation of time series, which is basically a sequence of symbols representing the raw numeric data. This low-dimensional form allows data mining algorithms to run efficiently, without negatively affecting their accuracy. 
However, in the area of time series classification (TSC), SAX-based classifiers have been unable to overtake traditional distance-based methods such as 1-Nearest Neighbour with Dynamic Time Warping (DTW) or Euclidean distance \cite{ding:vldb08,wang:dmkd13}. Nevertheless, these methods also have their shortcomings: they suffer from high running times and are negatively affected by noisy data. 
It is also not possible to interpret their classification decision: we can compare the test time series to a labelled nearest neighbour, 
but we get no information about the important time series aspects that lead to the classification decision. 
On the other hand, approaches including shapelet-based~\cite{grabocka-lts:kdd14,biST2015}, ensemble-based~\cite{bagnall2015time}, and more recently, deep learning algorithms~\cite{zwang.dnn}, 
have attracted more attention with their remarkable accuracy results. While SAX seemed to reach its limits regarding accuracy, 
another symbolic representation was introduced by Sch\"{a}fer et al.~\cite{Schafer:2012:SSF:2247596.2247656} and 
quickly became a strong competitor in this field. The new representation, named Symbolic Fourier Approximation (SFA), 
and its classification frameworks (BOSS, WEASEL and MUSE) \cite{schafer2015boss,schaefer:dmkd16,Schafer:2017:weasel,muse}, have further advanced the state-of-the-art regarding both accuracy and efficiency. 
These latter classifiers are the most accurate among dictionary-based algorithms, 
and are also comparable in accuracy to more complex classifiers based on large ensembles, e.g., COTE \cite{bagnall2015time,bagnall2016great}, but have much faster running times.

The appealing properties of symbolic representations (low-dimensional and interpretable) and the reinvigorated interest in their applications for classification, are the motivations for our study. 
Previously, we have examined the capability of SAX and proposed a new interpretable time series classifier~\cite{fvseql:7930038}. 
In this work, we explore other ideas in order to exploit the power of this and other symbolic representation techniques.

A symbolic representation of time series data is a string-based sequential description of the time series. 
The set of symbols for the description is usually predefined. Finding a meaningful symbolic representation is not trivial.
Symbolic techniques usually rely on a set of fixed parameters which have a strong impact on the resulting representation and classifier accuracy.
Finding a meaningful symbolic representation can thus be translated to finding the corresponding set of parameters. 
In addition, evaluating a chosen representation is also problematic and in many cases it is done via the classification model built upon it.
We note that relying on a single symbolic representation for any analysis task may not be appropriate.
Any kind of transformation induces loss of information. Hence, even the globally best representation might not contain enough information for the intended task. 
The process of finding the optimal symbolic representation is also costly. The SAX-VSM algorithm \cite{senin-saxvsm:icdm13} is such an example: it attempts to find the optimal SAX transformation parameters 
with an optimization algorithm (DIRECT) and an evaluation method (cross-validation). However, it fails to outperform the accuracy of recent state-of-the-art classifiers, e.g., BOSS, WEASEL, COTE.
On the other hand, using multiple representations have been considered and tested with some success. 
Classifiers such as Fast Shapelets \cite{fast-shapelet:sdm13}, BOSS \cite{schafer2015boss} and WEASEL \cite{Schafer:2017:weasel} 
generate symbolic representations at multiple resolutions by varying the symbolic representation parameters (e.g., working with 
multiple SFA representations obtained with different fixed parameters).
This shotgun approach, despite its simplicity, showed to be more robust and efficient than globally optimising the parameters of one fixed representation.
Another motivation for multiple representations of time series is the possibility of exploring different representation domains. 
Representations from different domains describe time series from different perspectives,
thus, by combining knowledge from different domains, it is possible to train a better model. 
An ensemble like COTE \cite{bagnall2015time} takes advantage of knowledge of time series as extracted from different domains. 
However, COTE gathers knowledge provided by its member classifiers, which have a combined time complexity that makes this approach impractical for 
many real-world problems.

This makes symbolic representations more appealing as they can standardize the representations in a sequential structure, 
effectively unifying knowledge from different domains without the need of multiple classification algorithms.
Existing approaches mostly combine symbolic representations at multiple resolutions (e.g.,  BOSS and WEASEL use multiple SFA resolutions), 
but do not take advantage of symbolic representations over multiple resolutions and multiple domains, possibly because this would generate a very large feature space.
The key challenge is to be able to efficiently work with a huge feature space of symbolic words from different representations, 
and ideally select the best feature subset without having to explicitly evaluate each feature.

In this work we propose a new classification framework for symbolic representation of time series. 
The core classifier of our framework is a symbolic sequence learning algorithm named SEQL~\cite{ifrim-seql:kdd11}, which was already shown to be suitable for the time series classification task~\cite{fvseql:7930038}. 
In this framework, we define the requirements for the representations and propose two different approaches for time series classification with multiple symbolic representations as input. 
The first approach is an ensemble of SEQL models trained using multiple symbolic representations.
The second approach uses SEQL as a feature selection method, combined with a linear learning algorithm (logistic regression).
We study SAX,  a representation in the time domain, and SFA, a representation in the frequency domain, as candidate representations for our methods. 
In our experiments, we explore different combinations for the input representation: a single symbolic representation (e.g., SAX with fixed parameters), 
multiple symbolic representations of the same type (e.g., multiple SAX resolutions) and multiple resolutions of symbolic representations of different types (e.g., multiple resolutions and multiple domains by combining SAX and SFA representations). 
We evaluate our algorithms on the well-known UCR time series classification benchmark \cite{ucr:keogh15} and discuss the impact on accuracy and efficiency 
of the different representations and algorithms on different types of problems and data types (e.g., motion, image, sensor, ECG). 

In this paper, we also investigate another aspect of TSC which is the interpretability of the model. The ability to explain a classification decision is often excluded from the discussion in this field, 
as the community has focused mostly on the classifier accuracy and efficiency. However, while accuracy and efficiency are very important, the model interpretability should be an essential evaluation concern for any time series classifier. 
In many applications, it is important to know the key characteristics of the data which are relevant for the analysis task or to understand the classification decision. 
For the TSC problem, we want to help users focus on the region that characterizes the data, i.e., the region examined by the model in order to make predictions. 
Our main interpretable classifier is a linear model, so we can use the weighted features learned by the model to emphasize the parts of the time series that lead to a classification decision.

\textbf{Our main contributions are as follows:}
\begin{itemize}
\item We present new TSC algorithms which can incorporate symbolic representations of multiple resolutions and multiple domains (SAX and SFA), using an efficient learning algorithm (SEQL).
\item We analyse the theoretical time complexity of all our proposed TSC algorithms.
\item We present an extensive experimental study of our approaches on the UCR Time Series Archive to understand the impact of multiple representations and compare to the state-of-the-art TSC methods.
\item We demonstrate how to interpret the linear classification models in the context of time series analysis.
\item We conduct a case study on the interpretability of our TSC models for an ecology benchmark. 
\end{itemize}




\section{Related Work}
\label{sec:relwork}


Traditionally, Euclidean and Time Warping distances are known to be very effective for time series analysis. 
Wang et al~\cite{wang:dmkd13} studied 8 different time series representations and 9 similarity measures and evaluated their performance on 38 datasets across various domains and tasks. 
In particular, for the TSC task, the study used 1-Nearest-Neighbor (1NN) classifiers to evaluate the accuracy of these measures. 
The conclusions provide interesting insights into the effectiveness of these measures and reaffirm the competitiveness of DTW in comparison to newer methods.

Recently, there has been also notable interest in shapelet-based classification algorithms after the first proposal by Ye et al. \cite{ye-shapelets:kdd09}. Shapelets are discriminative segments extracted from time series and can be used for classification. Moreover shapelets are interpretable, thus they can offer insight into the data. Since shapelet discovery is usually time-consuming, studies in \cite{Ye2011,fast-shapelet:sdm13,Gordon_shapelet} focused on enhancing the efficiency of the process. On the other hand, \cite{grabocka-lts:kdd14} formulated the problem as an optimization task and solved it with a stochastic gradient learning algorithm. The studies \cite{Lines:2012:STT:2339530.2339579,biST2015} used the shapelets to create a transformed dataset, in which the distance between the time series and a shapelet is a feature. 

COTE~\cite{bagnall2015time} (or Flat-COTE) is an ensemble method which incorporates 35 different classifiers for TSC.
HIVE-COTE~\cite{hive-cote}, a recent extension of COTE, added 3 more classifiers to the collection. 
Overall, COTE is among the most accurate TSC algorithms that have been tested on the UCR benchmark \cite{ucr:keogh15}. 
To the best of our knowledge, it is the only classifier that incorporates descriptions of time series from different domains. 
However, its learning framework is based on a large ensemble. This demands substantial computation resources as COTE's time complexity is determined by the slowest algorithm.

The excellent recent survey on TSC \cite{bagnall2016great} has contributed a systematic framework to evaluate time series classifiers.
In this study, the authors reproduced the experiments of 18 state-of-the-art classifiers in addition to two baseline classifiers (1-NN DTW and Rotation Forest)
 on the extended UCR benchmark that includes 85 datasets across a range of different TSC problem types (e.g., motion, image, ECG). 
The results were analysed based on algorithm type and problem type. The main claim of the survey was that benchmark classifiers are hard to beat and that COTE was by far the most accurate algorithm. 
Nevertheless, the authors mainly focused on the accuracy for evaluation, and did not evaluate either the efficiency or interpretability of the methods compared.


Regarding symbolic representations of time series, SAX is perhaps the most studied representation \cite{lin-sax:dmkd03,lin-sax:dmkd07,Lin2012:iis,fast-shapelet:sdm13}. For the classification task, SAX-VSM~\cite{senin-saxvsm:icdm13} is the most accurate approach to date among SAX-based methods. SAX-VSM builds a dictionary of SAX words from training data and computes a vector of tf-idf weights for each class. In addition, it employs an optimization algorithm to search for the optimal parameters of SAX in the parameter space. However the tuning cost is substantial due to the need of cross-validation. In addition, the SAX-VSM authors have analysed the interpretability of SAX-VSM models by mapping SAX words of high tf-idf scores back to the original time series. Our prior work ~\cite{fvseql:7930038} proposed a novel TSC algorithm using symbolic representations.
Our approach was a combination of a symbolic representation (SAX) and two adaptations of a sequence classifier (SEQL~\cite{ifrim-seql:kdd11}). 
Our proposed classifier, SAX-VFSEQL, can learn subsequences of symbolic words and can thus reduce the influence of SAX parameters and noise on the classification accuracy.
We have showed that SAX-VFSEQL delivers efficient and accurate models, while providing interpretability. Although more flexible than previous approaches that cannot learn symbolic sub-words,
 the accuracy of SAX-VFSEQL still suffers from being limited to only one fixed SAX representation (fixed set of parameters for the symbolic representation).

In \cite{Schafer:2012:SSF:2247596.2247656}, a new symbolic representation was introduced to index time series, the Symbolic Fourier Approximation (SFA). This approach 
uses a Discrete Fourier Transform as the core approximation technique. Based on this work, the authors proposed several classification frameworks for time series, 
which includes 1NN-BOSS  \cite{schafer2015boss}, BOSS VS  \cite{schaefer:dmkd16} (ensemble methods), WEASEL  \cite{Schafer:2017:weasel}, and MUSE \cite{muse}. 
In the SFA-based TSC algorithms family, WEASEL is the most recent work on univariate time series, while MUSE was developed to classify multivariate data. 
Both methods employ heavy feature engineering and selection techniques to filter the huge feature space created by multiple resolutions of SFA transformations, before feeding the 
selected features to a linear model. These approaches seem to suffer from efficiency issues since they do not include effective methods for pruning features early, so they need to 
carefully restrict the feature space (e.g., by restricting the SFA parameters and the type of features).
Nonetheless, according to the reported results, these methods are very fast and accurate, even when being compared to the ensemble classifiers (e.g., COTE). 
The authors never discuss the interpretability of these methods, arguably because of the non-linear characteristics of the SFA transformation.
 
The popularity surge of deep learning has inspired various studies to exploit its power for TSC. 
Recently the study of ~\cite{zwang.dnn} examined the performance of deep learning approaches on the UCR Archive. 
That work proposed 3 different Deep Learning architectures for TSC, namely Fully Convolutional Network (FCN), Multilayer Perceptrons (MLP), and Residual Network (ResNet), 
along with a new TSC evaluation measure, the Mean Per-Class Error (MPCE). 
For interpretability, the authors claimed that the FCN method is capable of identifying the regions of the time series contributing to the classification decision, by using a class activation map.
No detailed case study was conducted to substantiate the interpretability claim.
We compare our proposed TSC algorithms to these deep learning approaches and discuss their accuracy as reported in \cite{zwang.dnn}. Regarding measuring efficiency, 
this is not reported in that work. 
That work emphasizes the simplicity of the method and its effectiveness as a TSC baseline.
We downloaded the code but we could not reproduce the experiments on a regular PC in a reasonable amount of time (as we did with the other existing algorithms).
\section{Symbolic Representation of Time Series}
\label{sec:prelims}

In this section, we discuss two notable symbolic representations of time series: 
the Symbolic Aggregate approXimation (SAX) \cite{lin-sax:dmkd03} and the Symbolic Fourier Approximation (SFA) \cite{Schafer:2012:SSF:2247596.2247656}. 
Although they are different techniques that generate descriptions of time series in different domains, SAX and SFA produce very similar output in terms of structure. 
This property is strongly desirable in our approach as our classifiers require that the symbolic representation, regardless of the method, has a certain standard structure. 
Moreover, both have been showen to be powerful representations for the TSC task.

Generally, the output of both techniques can be described as a sequence of symbols taken from an alphabet $\alpha$, e.g., \textit{aaba}. 
In practice, it is common to employ a sliding window of length $l$ and repeatedly apply the transformation on the time series within this window. 
As a result, the output is a sequence of words, each of which is actually a symbolic sequence of length $w$, e.g., \textit{abba abbc bacc aacc}.

Formally, a symbolic sequence $S$ of length $n$ has the following form:

\begin{center}
$S$ = $s_1 s_2 \dots s_n$ where $s_i$ is in $\lbrace a_1,a_2, \dots a_{\alpha}\rbrace \cup \lbrace \textvisiblespace \rbrace$
\end{center}
The space character guarantees that S can be either a sequence of symbols or a sequence of words. We will show that the output of SAX and SFA can take the above form and therefore can work with our classifier. 
 Table~\ref{table:notation-tsc} summarises the notations used in this paper for the symbolic representations. 
\begin{table}[tbh]
\begin{center}
 \caption{Notation for the Time Series Classification (TSC) framework.}
  \label{table:notation-tsc}
\begin{tabular}{ll}\hline
Symbols & Description\\
\hline
$V$ & Raw (normalized) numeric time series\\
$N$ & Number of time series\\
$L$ & Length of original time series\\
$w$ & Length of a symbolic word \\
$\alpha$ & Size of alphabet \\
$l$ & Size of sliding window \\\hline
\end{tabular}
\end{center}
\end{table}

\subsection{Symbolic Aggregate approXimation}

SAX \cite{lin-sax:dmkd03} is a transformation method to convert a numeric sequence (time series) to a symbolic representation, 
i.e., a sequence of symbols with a predefined length $w$ and an alphabet of size $\alpha$. 
Generally, the technique includes three steps: 
\begin{enumerate}
\item Compute the Piecewise Aggregate Approximation (PAA)~\cite{PAA2001} of the time series.
\item Compute the lookup table for the given alphabet.
\item Map the PAA to a symbolic sequence by using the lookup table.
\end{enumerate}
\begin{figure}[!ht]
\centering
\includegraphics[width=0.8\textwidth]{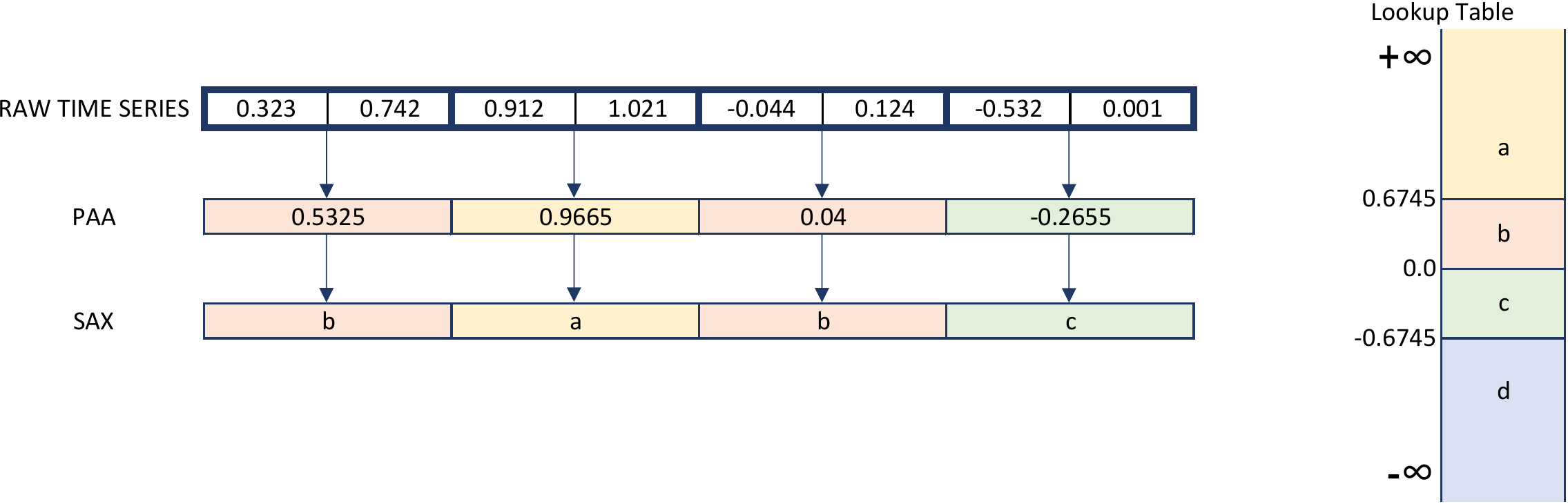}
\caption{An example of SAX transformation that includes: the raw time series, the Piecewise Aggregate Approximation, the lookup table and the final SAX output for $(w=4,\alpha=4)$ .}
\label{fig:saxchars}
\end{figure}

In the first step, the time series is z-normalized and then divided into $w$ equal-length segments and each segment is replaced with its mean value. The result is the PAA vector of length $w$. In the second step, a lookup table is built for the alphabet $\alpha$. Each symbol in the alphabet is associated with an interval, i.e., a continuous range of values. The intervals are obtained by dividing the domain of the time series to $\alpha$ disjoint areas with equal probability, assuming that the values of the time series are normally distributed (hence the z-normalization). Finally, each entry of the PAA vector is then replaced by a symbol taken from the alphabet by using this lookup table. 

Figure~\ref{fig:saxchars} illustrates an example of SAX output with parameters set to $L = 8$, $w = 4$ and $\alpha = 4$. The lookup table divides the domain of the time series into 4 intervals by defining 3 breakpoints ($-0.6745$, $0.0$, and $0.6745$) and links each interval to a symbol from the alphabet $\{a,b,c,d\}$. Each entry in the PAA vector is the average of the corresponding segment in the raw time series. The SAX sequence is produced by looking up the PAA from the table. The first entry ($0.5325$) falls within the range $[0.0,0.06745)$ thus the first symbol taken is $b$.

\begin{figure}[!ht]
\centering
\includegraphics[width=0.8\textwidth]{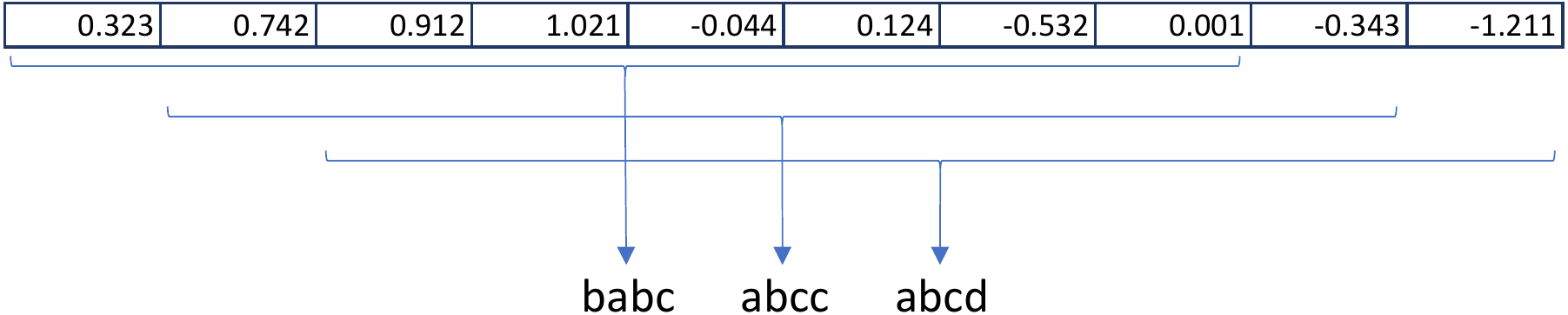}
\caption{Sliding window shifting to obtain a time series representation based on SAX-words $(l=8,w=4,\alpha=4)$ .} 
\label{fig:saxwords}
\end{figure}

SAX can also be combined with a sliding window of length $l$, usually done to process longer time series (Figure~\ref{fig:saxwords}). 
Our previous study \cite{fvseql:7930038} also found that the sliding window technique has a positive impact on the classification accuracy, 
arguably because it can capture a finer description of the time series.

The procedure to transform a time series to a SAX representation with a sliding window can be summarised in Algorithm~\ref{alg:sax_with_sliding_window}. 
The sliding window starts from the first time-stamp, i.e., the beginning of the time series. 
The subsequence of length $l$ within the window is then transformed to a symbolic sequence of length $w$ with the described above steps. 
This sequence is commonly referred to as a SAX word. The process is repeated until the window reaches the end of the time series. 
Hence the final output is a sequence of equal-length SAX words.

\begin{algorithm}[h]
\caption{SAX with sliding window}
\begin{algorithmic}[1]
\State Set window length $l$
\State Set word size $w$
\State Set alphabet size $\alpha$
\State Compute lookuptable
\State $L = length(timeseries)$
\ForAll{$t$ in $\left[ 0 , L - l \right) $}
\State $normed\_ts = z\_normalize(timeseries[t:t+l])$
\State $PAA = computePAA(normed\_ts)$
\State $S = ""$
\For{$v$ in PAA}
\State $S \mathrel{+}= lookup(v)$
\EndFor
\State Add $S$ to the final representation
\EndFor
\end{algorithmic}
\label{alg:sax_with_sliding_window}
\end{algorithm}

\subsection{Symbolic Fourier Approximation}

SFA \cite{Schafer:2012:SSF:2247596.2247656} also transforms a time series to a symbolic representation. Similarly to SAX, SFA employs a sliding window to extract segments of time series before transformation (although the authors of \cite{Schafer:2017:weasel} also suggested non-overlapping window for their SFA-based WEASEL classification framework). Hence SFA's parameters also include the window size $l$, the word length $w$ and the alphabet size $\alpha$. 

The core differences between SAX and SFA are the choices of approximation and discretisation techniques. SFA uses a Discreet Fourier Transform (DFT) method to approximate a time series. 
DFT is well known in the signal processing community and can act as a filter to remove noise from data. 
The same authors also introduced a Multiple Coefficient Binning (MCB) method to discretise the approximation (i.e., a vector of length $w$). 
The overall procedure consists of two major steps:

\begin{enumerate}
\item MCB discretisation: Compute the lookup table from the DFT approximations of the training data.
\item Map the DFT approximation of the input time series to its SFA representation with the lookup table.
\end{enumerate}

Both steps employ the DFT technique to approximate an input time series with a vector of length $w$. Basically, DFT decomposes a time series into a series of sinusoid waves, 
each of which can be represented by a Fourier coefficient. The coefficient is a complex number, hence it can be defined by 2 real values: one for the imaginary part and one for the real part. 
Therefore, only the first $w/2$ coefficients of the series are used to create a sequence of length $w$.

\begin{figure}[!ht]
\centering
\includegraphics[width=0.8\textwidth]{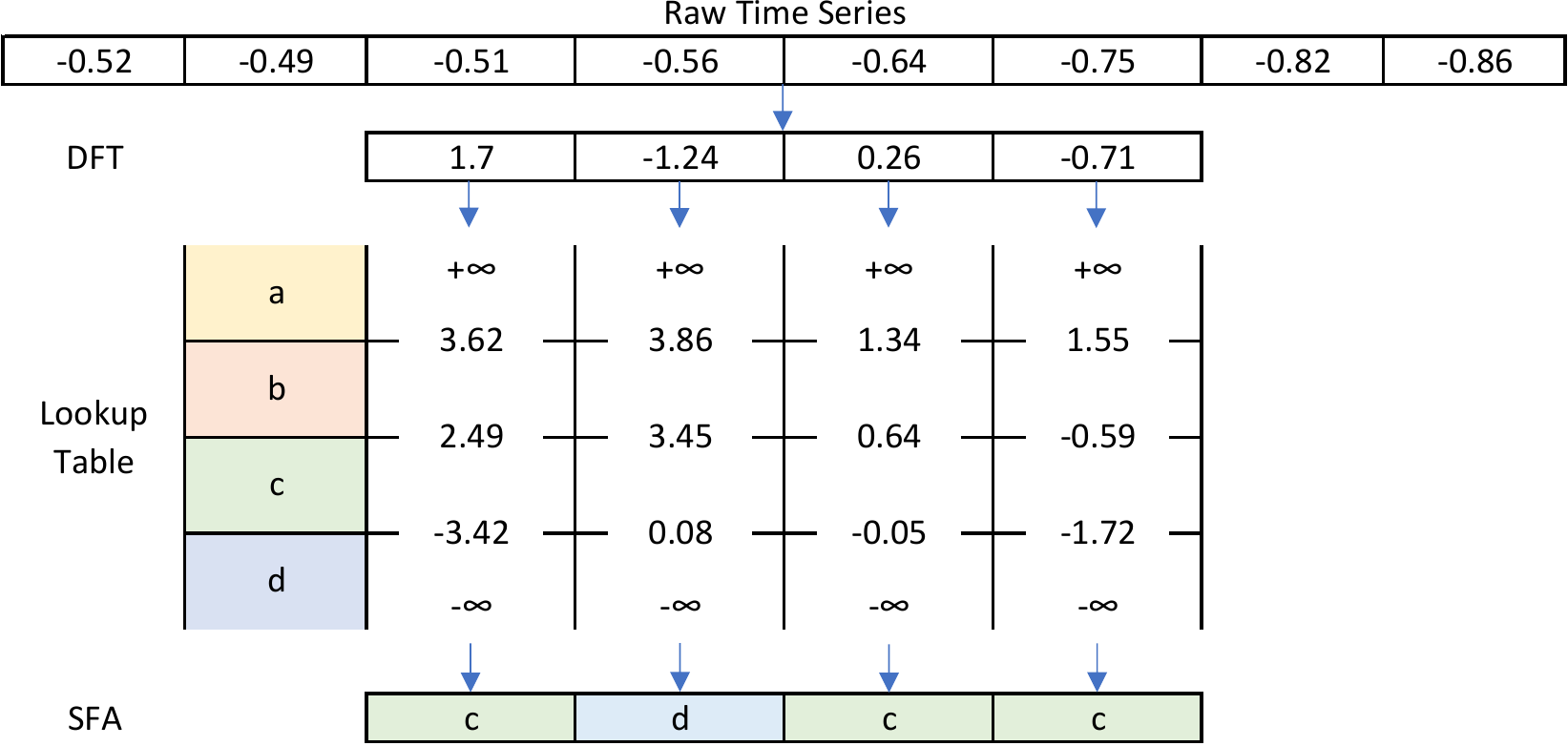}
\caption{An example of SFA transformation that includes: the raw time series, the DFT approximation, the lookup table and the final SFA output for $(w=4,\alpha=4)$ .}
\label{fig:sfa-example}
\end{figure}

The discretisation step (MCB) computes the lookup table from the training data. It first computes the DFT approximations of all $l$-length segments extracted from the entire training data to obtain a set of $w$-length vectors. Then from this set of vectors, it computes a set of $\alpha - 1$ breakpoints for each i-th position ($1 \leq i \leq w$) according to the distribution of the i-th values (i.e., equi-depth binning method). The areas divided by the breakpoints are associated with the symbols from the alphabet. The result is a lookup table of $w$ columns and $\alpha$ rows (Figure~\ref{fig:sfa-example}).

The transformation step first computes the DFT approximation for each $l$-length segment of the input time series. For each i-th entry of the approximation vector, the corresponding symbol is looked up from the i-th column of the lookup table. Thus each segment of the time series is transformed to a sequence of length $w$. The final result is a sequence of SFA words extracted from the input time series.

Figure~\ref{fig:sfa-example} gives an example of SFA transformation. The raw time series is DFT approximated with a vector of length $w=4$. The i-th entry of this vector is then replaced by a symbol according to the i-th column in the lookup table.
Table~\ref{table:sax_vs_sfa} summarizes the difference between the SAX and SFA representations. The time complexity is reported from \cite{Schafer:2012:SSF:2247596.2247656}.

\begin{table}[!ht]
\begin{center}
\caption{Comparison between SAX and SFA symbolic representations of time series.}
\label{table:sax_vs_sfa}
\begin{tabular}{|c|c|c|c|}
\hline 
Method & Approximation & Discretisation & Complexity \\ 
\hline 
SAX & PAA & equi-prob & $\mathcal{O}(NL\log{}L)$ \\ 
\hline 
SFA & DFT & MCB + equi-depth  & $\mathcal{O}(NL)$ \\ 
\hline 
\end{tabular} 
\end{center}
\end{table}

\section{Sequence Learner with Multiple Symbolic Representations of Time Series}
\label{sec:method}

Typically, the SAX representation is susceptible to how we select parameters, 
i.e., the window size, the length of the sequence or the size of the alphabet. 
Each choice of parameters captures a different structure of the time series which is essential for the classification task. 
One solution for this issue is to search for the optimal parameters, either by a naive grid search or a more complex optimization algorithm
 (e.g., DIRECT as in SAX-VSM \cite{senin-saxvsm:icdm13}).  In our previous work, we mitigate this issue by introducing a new algorithm that can learn discriminative sub-words from a SAX word-based representation~\cite{fvseql:7930038}.

To improve our results, we study here an alternative approach which uses multiple resolutions and multiple domain representations of time series. 
Fast Shapelets, BOSS and WEASEL are notable classifiers using multiple resolutions, although they only support representations from one domain. 
Their competitive results (in terms of both accuracy and efficiency) demonstrates the potential of this approach. 
We hypothesize that, a single representation of time series, even an optimal one, might be insufficient to capture the necessary structure for the classification task. 
By combining knowledge from multiple resolutions and representations, a learning algorithm can deliver a more robust model.

This section first describes the core technique of our approach, which is a classification framework for the (single) symbolic representation of time series. After that, 
two different methods based on the core technique are proposed, to make uses of multiple representations. The first one is an ensemble method and the second one is a feature selection method.

\subsection{Sequence Learner with Symbolic Representation of Time Series}

\begin{figure}[!ht]
\centering
\includegraphics[width=0.7\textwidth]{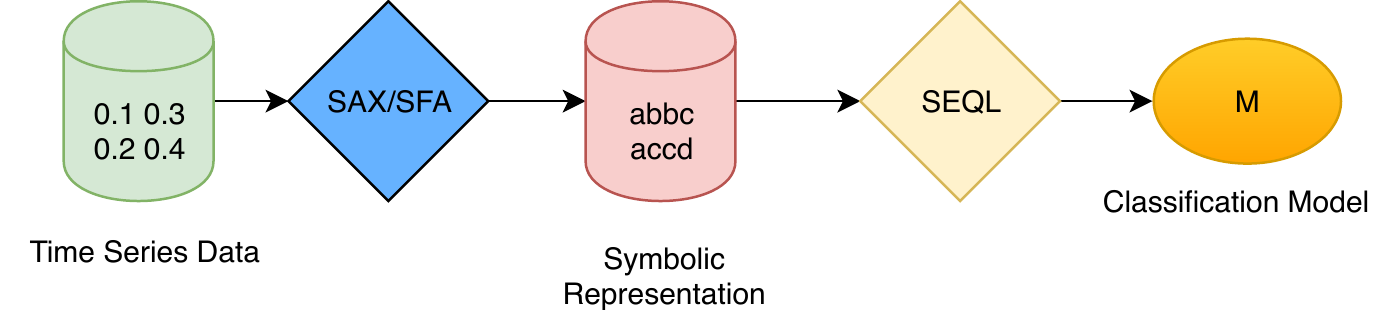}
\caption{SEQL with symbolic representation (SAX or SFA) of time series.} 
\label{fig:sax_vseql}
\end{figure}

Figure~\ref{fig:sax_vseql} sketches our approach which is composed of two components: a symbolic representation and a classifier. 
The symbolic representation can be either SAX or SFA, which were already discussed in the previous section. 

Our core algorithm for classification is Sequence Learner (SEQL) \cite{ifrim-seql:kdd11}. In~\cite{fvseql:7930038}, two adaptations of SEQL were introduced for the TSC task. 
The first one (SAX-VSEQL) can learn subsequences from the SAX words while the second one (SAX-VFSEQL) can approximate the subsequences. 
The latter was proposed mainly to make the representation less dependent on the symbolic parameters ($l$, $w$, and $\alpha$). 
As multiple resolutions of a given symbolic representation can create the same effect, we decided to use only the lightweight VSEQL version as the core classifier for our experiments.
Hence from here on, we use the name SEQL to refer to the VSEQL TSC classifier from \cite{fvseql:7930038}.

\paragraph{\textbf{SEQL}} was originally designed as a binary classifier for sequence data such as DNA or text. 
The algorithm is  able to explore the all-subsequence space by employing a branch-and-bound feature search strategy. 
Thus it can find a set of discriminative subsequences in an effective manner. With the symbolic representation of time series, 
the results can easily be translated to a set of time series' discriminative segments. TSC with symbolic representation is not a new idea, 
however the most common approach is to build a dictionary directly from the results of the transformation (usually a bag of symbolic words). 
In~\cite{fvseql:7930038}, we showed that SEQL is suitable for the TSC task, albeit some adaptations are required to learn discriminative sub-words.

The training input for SEQL is a set of sequences and their labels. 
The output is a linear mapping function $f: S\rightarrow  \{-1, +1\}$ to predict the label of unseen data. 
The function $f$ is represented by a parameter vector $\beta = (\beta_1, \dots, \beta_j,\dots, \beta_d)$ which minimizes the loss function $L(\beta)$ using greedy coordinate descent:

\begin{equation}
\label{eq:minL}
\beta^{*}=\argmin_{\beta \in \mathbb{R}^{d}}L(\beta)
\end{equation}
where
\begin{equation}
\label{eq:generalLoss}
L(\beta) = \sum_{i=1}^{N} \xi(y_i,x_i,\beta)+ CR_\alpha (\beta)
\end{equation}
and $\xi(y_i,x_i,\beta)$ is the binomial log-likelihood loss:
\begin{equation}
\label{eq:log-loss}
\xi(y_i,x_i,\beta) = log(1+e^{-y_i\beta^{T} x_i})
\end{equation}

The $x_i$ in Equation~\ref{eq:generalLoss} and \ref{eq:log-loss} denotes the feature vector of the sequence $S_i$ (in all-subsequence feature space) while $y_i$ denotes its true label (either $+1$ or $-1$). On the right hand side of Equation~\ref{eq:generalLoss}, $C$ denotes the weight and $R_\alpha (\beta)$ denotes the elastic-net regularization. In SEQL the all-subsquence feature space is not explicitly generated, rather an iterative process is applied to efficiently 
search the feature space for the best feature to be optimized next. The classification decision is computed as $f(x) = \beta^{t} x$. 

\begin{algorithm}[h]
\caption{SEQL workflow}
\label{alg:BasicWorkflow}
\begin{algorithmic}[1]
\State Set $\beta^{(0)}=0$
\State Find all unigrams and their positions in the training data $seeds = allUnigrams()$
\While{!termination condition}
\State Calculate objective function $L(\beta^{(t)})$
\State Find best subsequence with maximum gradient value ${\hat{s}_j}$ = findBestNGram(seeds) //\emph{Algorithm \ref{alg:tree_expansion} for Tree Search}
\State Find optimal step length $\eta^{(t)}$ \
\State Update $ \bm{\beta}^{(t)}=\bm{\beta}^{(t-1)}-\eta^{(t)}\frac{\bm{\partial}L}{\bm{\partial}\beta^{(t)}_{j}}(\bm{\beta}^{(t-1)})$
\State Add subsequence $\hat{s}_j$ to feature set
\EndWhile
\end{algorithmic}
\label{alg:seq}
\end{algorithm}

The SEQL workflow is shown in Algorithm~\ref{alg:seq}. It iteratively optimizes $\beta$ to minimize the loss in Equation~\ref{eq:generalLoss}. At each iteration, it selects the most discriminative subsequence (Line 5) using Gauss-Southwell selection and optimizes its corresponding coefficient $\beta_i$ (Line 7). To select the most discriminative subsequence, it navigates a sequence tree in which each node is a subsequence in the all-subsequence space. 

Algorithm~\ref{alg:tree_expansion} portrays the breadth-first-search algorithm used by SEQL: Each $candidate$ is a node in the tree and each node stores all the locations of the corresponding subsequence in the training data (Line 7). Line 9 guarantees that the expansion is restricted within a word of the symbolic representation. The unigrams are the single characters in the symbolic alphabet.

\begin{algorithm}[h]
\caption{(V)SEQL: Tree Expansion and Search}
\begin{algorithmic}[1]
\Function{findBestNGram}{seeds}
\State $candidates = seeds.duplicate()$
\State $new\_candidates = \lbrace\rbrace$
\While{$candidates$ is not empty}
\ForAll{$candidate$ in $candidates$}
\If{not $can\_prune(candidate)$}
\ForAll{$loc$ in $candidate \rightarrow locations$}
\State $next\_symbol = symbol\_at(loc+1)$
\If{$next\_symbol$ is not space}
\State Add new Node($candidate + next\_symbol$) to $new\_candidates$
\EndIf
\EndFor
\If{$is\_best\_candidate(candidate)$}
\State Update best candidate
\EndIf
\EndIf
\EndFor
\State $candidates = new\_candidates$
\EndWhile
\State Return best candidate
\EndFunction
\end{algorithmic}
\label{alg:tree_expansion}
\end{algorithm}

The size of a complete tree is the total number of subsequences, which is normally impractical to traverse. However, SEQL employs a greedy branch-and-bound strategy to effectively prune any unpromising part of the tree (Line 6). This strategy relies on the anti-monotonicity property of a sequence, i.e., a sequence is always equally or less frequent than all of its subsequences. This property allows the algorithm to calculate a gradient upper-bound for each subsequence, which effectively makes the pruning decision. More details about the algorithm can be found in the original paper~\cite{ifrim-seql:kdd11}.

The output of SEQL training is a linear model which is essentially a set of subsequences and their coefficients (Table~\ref{table:seql_model}). The coefficients can be interpreted as the discriminative power of the subsequence.

\begin{table}[htb]
    \sisetup{round-mode=places, round-precision=3}
    \centering
    \caption{An example model trained by SEQL: a linear model in the space of all symbolic sub-sequences.}
   	\label{table:seql_model}
    \begin{filecontents*}{csv/coffee_seql_model.csv}
coef,subseq
0.0140603,cccdbcdda
0.0129099,cccdcbcddaaa
0.0120596,ccdbc
0.00716273,cccdbc
0.00435397,ccccdbc
-0.00604542,cccbbdd
-0.010253,cccbbd
-0.0105644,ccbbddb
-0.0122196,cccbb
-0.0142573,ccccbb
\end{filecontents*}
\csvreader[
               table head=\toprule {Coefficients} & {Subsequences}\\ \midrule,
               tabular={ll},
               head to column names,
               late after last line=\\\bottomrule]
               {csv/coffee_seql_model.csv}{}%
     {\coef & \subseq}%
\end{table}

\paragraph{\textbf{SAX-SEQL:}} The combination of SAX and SEQL is trivial (Figure~\ref{fig:sax_vseql}): the raw time series data for training is transformed to its SAX representation. SEQL learns a classification linear model from the new training data. The test data is also transformed to its SAX representation before being classified (Algorithm~\ref{alg:sax_seql}).

\begin{algorithm}[h]
\caption{The workflow of SAX-SEQL}
\begin{algorithmic}[1]
\State Set $l,w$ and $\alpha$
\State $train = SAXtransform(train\_time\_series, l,w,\alpha)$
\State $test = SAXtransform(test\_time\_series, l,w,\alpha)$
\State Train with SEQL $M = SEQLearner(train)$
\State Test with SEQL $predictions = SEQLClassifier(M,test)$
\end{algorithmic}
\label{alg:sax_seql}
\end{algorithm}



\paragraph{\textbf{SFA-SEQL:}}The original SFA representation has a tricky property which is in principle unsuitable for SEQL: the i-th symbol is always corresponding to the i-th column of the lookup table, thus the same symbol may not represent the same range of values. For example, the lookup table in Figure~\ref{fig:sfa-example} can produce a sequence ``$dabb$" in which the 3-rd and 4-th positions share the same symbol ``$b$". However, while the former represents the interval $[0.64,1.34)$, the latter corresponds to $[-0.59,1.55)$. Moreover, they are linked to the real part and the imaginary part of a Fourier coefficient. Unfortunately, SEQL would be misled by treating both ``$b$" as the same unigram. To fix this issue, we simply let each column in the lookup table have its own alphabet, e.g., the symbolic representation of the same example would be ``$d_{1}a_{2}b_{3}b_{4}$". In this way, SEQL can recognize two different unigrams ``$b_{3}$" and ``$b_{4}$". Technically, the size of the new alphabet is $w \times \alpha$ but we will still refer to the size of alphabet as $\alpha$ for the sake of simplicity.
Besides the above issue, the procedure for training and testing SFA-SEQL is identical to that of SAX-SEQL.

\subsection{Ensemble SEQL}

\begin{figure}[!ht]
\centering
\includegraphics[width=0.7\textwidth]{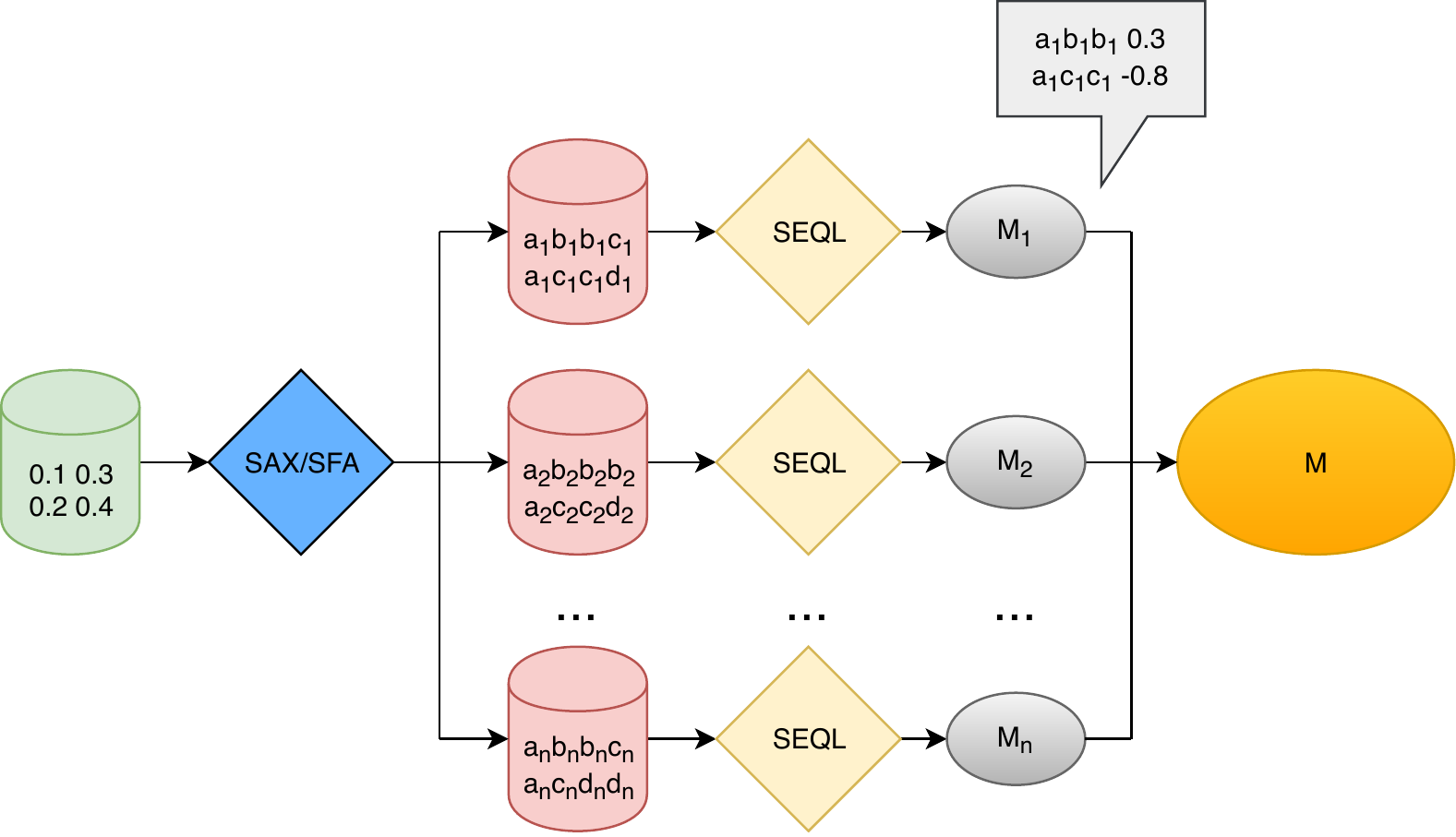}
\caption{An ensemble of SEQL models trained from different SAX (or SFA) representations. We call this a multi-resolution approach for a given symbolic representation.} 
\label{fig:ensemble_vseql}
\end{figure}

Ensemble SEQL, as the name suggests, is an ensemble of multiple SEQL models. Each model is trained by the same classifier (SEQL) but with a different symbolic representation of the training data as input. Figure~\ref{fig:ensemble_vseql} illustrates the training procedure to produce $n$ SEQL models from different SAX representations for the ensemble.
Different representations of the time series can be generated simply by adjusting the transformation parameters ($l$, $w$, and $\alpha$) to obtain multiple resolutions for a given symbolic representation. 
Algorithm~\ref{alg:ensemble_train} shows how we can adjust the window length $l$ to train multiple models (Line 3). As a result, the number of representations is approximately $sqrt(L)$, i.e., the longer the time series, the larger the number of representations it can produce. For each set of parameters, the raw training data is transformed to the the SAX representation (Line 4) and a new SEQL model $M_i$ is trained upon this representation (Line 5). The output is an ensemble $M$ of all $M_i$.

\begin{algorithm}[h]
\caption{Ensemble SEQL: Training}
\begin{algorithmic}[1]
\State Set word size $w=16$
\State Set alphabet size $\alpha=4$
\For{$l = minl$,$w <= L$, $w \mathrel{+}= sqrt(L)$}
\State $sax = SAXtransform(raw\_time\_series, l,w,\alpha)$
\State Train the SEQL model from the symbolic representation $M_i = SEQL(sax)$
\State $M[l,w,\alpha] = M_i$
\EndFor
\end{algorithmic}
\label{alg:ensemble_train}
\end{algorithm}

\begin{algorithm}[h]
\caption{Ensemble SEQL: Testing}
\begin{algorithmic}[1]
\State Set word size $w=16$
\State Set alphabet size $\alpha=4$
\State $score = 0$
\For{$l = minl$,$w <= L$, $w \mathrel{+}= sqrt(L)$}
\State $sax = SAXtransform(raw\_time\_series, l,w,\alpha)$
\State $score \mathrel{+}= M[l,w,\alpha].predict(sax)$
\EndFor
\State $prediction = sign(score)$
\end{algorithmic}
\label{alg:ensemble_test}
\end{algorithm}

For the prediction (Algorithm~\ref{alg:ensemble_test}), the unlabelled time series is converted to a SAX representation with the same set of configurations chosen in the training step (Line 5). 
Each model makes a prediction based on the representation of the corresponding configuration (Line 6). The sign of the predicted score aggregation will determine the predicted class of the time series (Line 8).


\subsection{SEQL as Feature Selection}

\begin{figure}[!ht]
\centering
\includegraphics[width=0.7\textwidth]{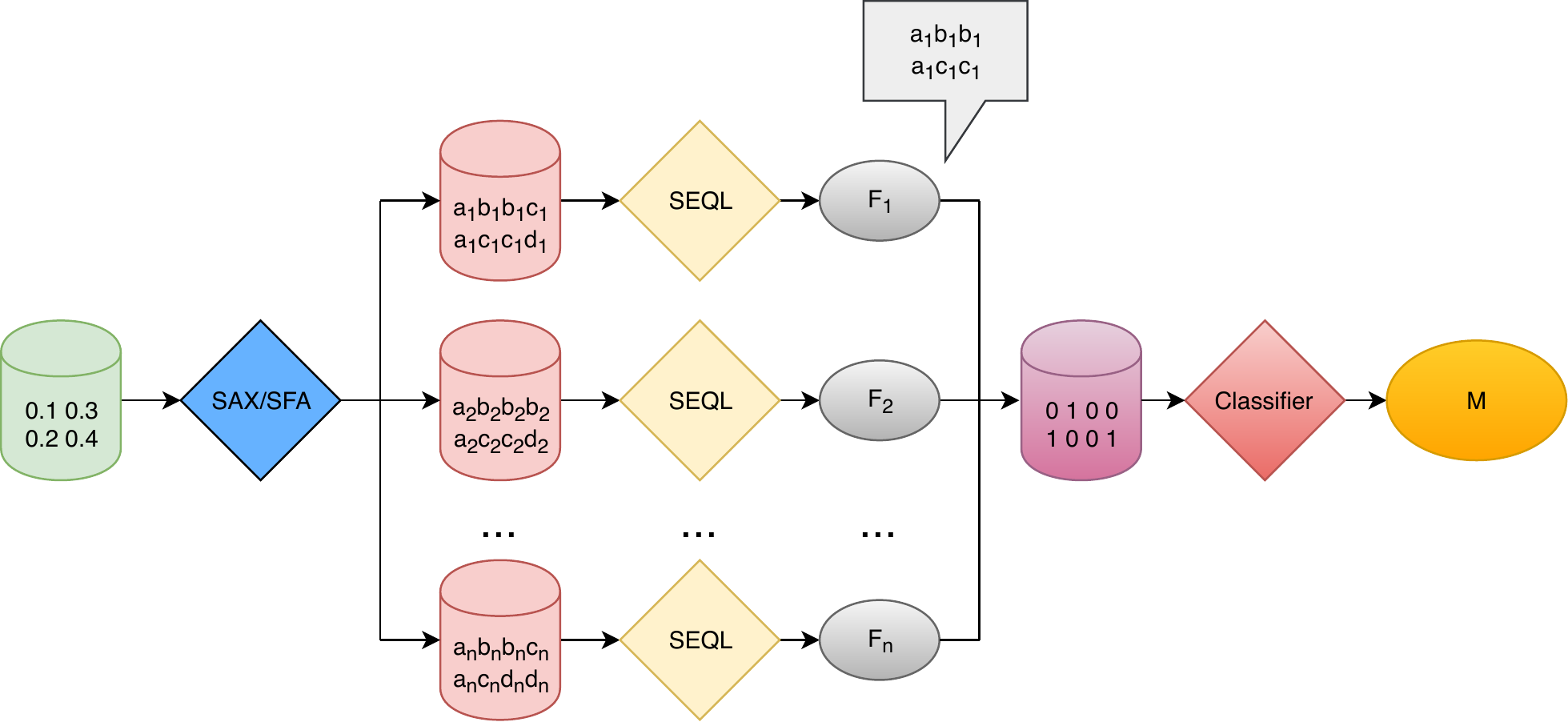}
\caption{SEQL as feature selection method. Features are selected from multiple-resolutions and/or multiple-domain symbolic representations and fed to a logistic regression algorithm.} 
\label{fig:feature_selection_vseql}
\end{figure}

The learning output of SEQL is essentially a list of sub-sequences selected from training data (Table~\ref{table:seql_model}), hence the method can be used for feature selection. The process diagram for this scheme is illustrated in Figure~\ref{fig:feature_selection_vseql}. 

As it can be seen from the diagram, we also first transform time series data to multiple symbolic representations. Again, the representations can be either SAX or SFA. We then feed each representation to a SEQL trainer. Algorithm~\ref{alg:seql_as_fs} explains how the features were extracted using SEQL. It is basically Algorithm~\ref{alg:ensemble_train} with an additional for-loop to collect the new features. 
A feature is identified by a subsequence learned by SEQL and the associated SAX (or SFA) configuration (Line 8).
With the collected set of features $F$, it is possible to apply any traditional classifying technique such as logistic regression, support vector machine or random forest. 
After experimenting with several learning algorithms, we chose logistic regression for its simplicity, accuracy and interpretability.

\begin{algorithm}[h]
\caption{SEQL as Feature Selection}
\begin{algorithmic}[1]
\State Set word size $w=16$
\State Set alphabet size $\alpha=4$
\State Set of features $F=\{\}$
\For{$l = minl$,$w <= L$, $w \mathrel{+}= sqrt(L)$}
\State $sax = SAXtransform(raw\_time\_series, l,w,a)$
\State Train the SEQL model $M_i = SEQL(sax)$
\ForAll{$subsequence$ in $M_i$}
\State $F$.add(new Feature($subsequence,l,w,\alpha$))
\EndFor
\EndFor
\end{algorithmic}
\label{alg:seql_as_fs}
\end{algorithm}

It is also worth to note that feature engineering methods are also applicable here. In fact, multiple representations can lead to strongly correlated features since they are essentially generated from the same time series (with different transformations). An extra step to filter unnecessary features might be useful in practice. However, we will demonstrate in our experiments that, even with the absence of such filters, the model learnt upon the SEQL feature selection is still very accurate on test data.

\subsection{SEQL with Multiple Representations from Multiple Domains}

So far we have discussed the algorithms in the scenario of single-type representations, i.e., either SAX or SFA. However, as SEQL can work with both representations, making use of both for classification is fairly straightforward. The ensemble $M$ can contain both SAX and SFA models: $M = M_{SAX} \cup M_{SFA}$. The set of features $F$ can also contain both SAX and SFA features: $F = F_{SAX} \cup F_{SFA}$.

Table \ref{table:exp_combi} summarizes different combinations between the symbolic representations and all proposed variants of SEQL-based algorithms.

\begin{table}[h]
\begin{center}
\caption{Combinations of symbolic representations and variants of SEQL-based algorithms.}
\label{table:exp_combi}
\begin{tabular}{|c|c|c|c|}
\hline 
Type & Symbolic Representation & SEQL variant & Name \\ 
\hline 
Single & SAX & SEQL & SAX-SEQL \\ 
\hline 
Single & SFA & SEQL & SFA-SEQL \\ 
\hline 
Multiple & SAX & Ensemble SEQL & mtSAX-SEQL \\ 
\hline 
Multiple & SFA & Ensemble SEQL & mtSFA-SEQL \\ 
\hline 
Multiple & SAX and SFA & Ensemble SEQL & mtSS-SEQL \\ 
\hline 
Multiple & SAX & SEQL as Feature Selection & mtSAX-SEQL+LR \\ 
\hline 
Multiple & SFA & SEQL as Feature Selection & mtSFA-SEQL+LR \\ 
\hline 
Multiple & SAX and SFA & SEQL as Feature Selection & mtSS-SEQL+LR \\ 
\hline 
\end{tabular} 
\end{center}
\end{table}



\subsection{Time Complexity}

The time complexity of SEQL is proportional to the number of the subsequences it has to evaluate. In the worst case, SEQL has to explore the complete subsequence space, i.e., when it fails to prune any part of the tree:

\begin{align}
T(\text{SEQL})
		& = \mathcal{O}(N_s(1 + \dots + l_s)) \nonumber \\
		& = \mathcal{O}(N_s l_s^2) \label{eq:seql_worst} \
\end{align}

where $N_s$ is the total number of sequences and $l_s$ is the length of the sequence.

For the symbolic representation of time series, each word is counted as a sequence, accordingly $N_s \leq N(L - l)$ and $l_s = w$:

\begin{align}
T(\text{SAX-SEQL})
		& = T(\text{SAX}) + T(\text{SEQL}) \nonumber \\
		& = \mathcal{O}(NL\log{}L) + \mathcal{O}(N (L-l) w^2) \nonumber \\
		& = \mathcal{O}(NL\log{}L) + \mathcal{O}(N L w^2) \label{eq:sax_seql_1} \
\end{align}

In practice, $w$ is often a constant, hence:

\begin{align}
T(\text{SAX-SEQL})
		& = \mathcal{O}(NL\log{}L) + \mathcal{O}(N L) \nonumber \\
		& = \mathcal{O}(NL\log{}L) \label{eq:sax_seql_2} \
\end{align}

If multiple representations are used, the complexity also depends on the number of representations, which is approximately $\sqrt{L}$:

\begin{equation}
\label{eq:mtsax_seql}
T(\text{mtSAX-SEQL}) = \mathcal{O}(NL^{\frac{3}{2}}\log{}L)
\end{equation}

Similarly, if the type of the representation is SFA:

\begin{equation}
\label{eq:sfa_seql}
T(\text{SFA-SEQL}) = \mathcal{O}(NL)
\end{equation}

\begin{equation}
\label{eq:mtsfa_seql}
T(\text{mtSFA-SEQL}) = \mathcal{O}(NL^{\frac{3}{2}})
\end{equation}

In practice, the running time can be reduced greatly due to several factors. When using symbolic transformations, repeated words can be discarded (often referred to as numerosity reduction in related literature \cite{lin-sax:dmkd03}). 
The pruning technique in SEQL is effective in practice and drastically diminishes the number of subsequences to be evaluated. Finally, parallelism is possible in the case of multiple representations since each training step is independent from 
other representations.
\section{Experimental Results}

In this paper we study two symbolic representations (i.e., SAX and SFA) and suggest different variants of SEQL-based algorithms for the TSC task. 
To test our approaches, we experiment in total with 8 different combinations between the two symbolic representations and the SEQL variants (Table~\ref{table:exp_combi}).
The parameter settings for the experiments are informed by existing literature on symbolic representations for TSC \cite{Schafer:2017:weasel,fvseql:7930038} and are set as in Table~\ref{table:params}. 
For multiple representations we use a minimum sliding window of size $l=20$ and an increment step of $\sqrt{L}$. For SEQL we use default parameters as in \cite{ifrim-seql:kdd11}.


\begin{table}[h]
\label{table:params}
\begin{center}
\caption{Parameter settings for the experiments: $l$ is the window size, $w$ is the word size and $\alpha$ is the alphabet size.}
\begin{tabular}{|c|c|c|c|}
\hline 
Input representation(s) & $l$ & $w$ & $\alpha$ \\ 
\hline 
Single SAX & $0.2*L$ & 16 & 4 \\ 
\hline 
Single SFA & $0.2*L$ & 8 & 4 \\ 
\hline 
Multiple SAX & varied & 16 & 4 \\ 
\hline 
Multiple SFA & varied & 8 & 4 \\ 
\hline 
\end{tabular} 
\end{center}
\end{table}

\subsection{Experiment Setup}



Our proposed algorithms were tested on all 85 datasets of the UCR Time Series Classification Archive~\cite{ucr:keogh15}. The archive has been incrementally extended by researchers working with time series and possesses a vast collection of data from multiple domains. It is perhaps the most common used benchmark for recent studies on TSC. 
Our test system is a Linux PC with Intel Core i7-4790 Processor (Quad Core HT, 3.60GHz), 16GB 1600 MHz memory and 256 Gb SSD storage. All our code was developed in C\nolinebreak\hspace{-.05em}\raisebox{.4ex}{\tiny\bf +}\nolinebreak\hspace{-.10em}\raisebox{.4ex}{\tiny\bf +} and can be found at \url{https://github.com/lnthach/Mr-SEQL}.

For comparison with state-of-the-art approaches, we take results directly from the recent survey \cite{bagnall2016great} or from the respective publications where 
each of the algorithms was originally published.

\subsection{Accuracy}

\begin{figure}[!ht]
\centering
\includegraphics[width=1.05\textwidth]{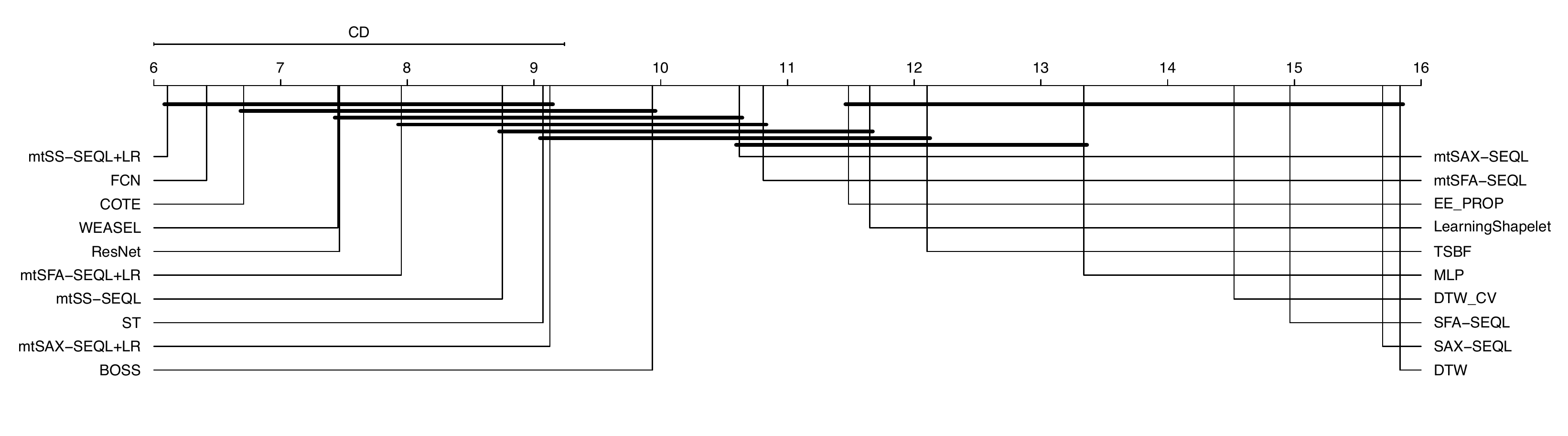}
\caption{Average error ranking of compared classifiers ordered (left-best) based on performance on the UCR Archive.}
\label{fig:sota1}
\end{figure}

Figure~\ref{fig:sota1} is a critical diagram that presents the ranking of the state-of-the-art classifiers including our proposed algorithms, based on the error attained on the UCR Archive.
This type of diagram is used very often in the literature on TSC for visual comparison of many different methods across many different datasets \cite{bagnall2016great,schafer2015boss}. 
The diagram shows the classifiers on a spectrum of average error ranking (the rank of the method with regard to classification error, averaged across 85 datasets), 
therefore classifiers to the left of the diagram perform better than classifiers to the right. The thick horizontal black line groups methods that do not have a statistically significant difference in performance.

Generally, the top rank accuracy group (on the left hand side) is composed of representatives from mtSS-SEQL+LR (linear classifier with SAX and SFA features), ensemble algorithms (COTE), deep learning algorithms (FCN and ResNet) and SFA-based classifiers (our proposal mtSFA-SEQL+LR and an existing approach named WEASEL). Our proposed algorithm, mtSS-SEQL+LR, tops the chart and has a similar average rank to COTE and FCN.
The middle rank group consists also of more ensemble representatives (EE\_PROP, mtSAX-SEQL and mtSFA-SEQL). The remaining ones to the right of the spectrum include the traditional method DTW and single representation methods (SAX-SEQL, SFA-SEQL).

Table~\ref{table:sota_stats} reports the average error and other performance measurements for the same classifiers. The Mean Per-Class Error  (MCPE) is an alternative performance measurement proposed by \cite{zwang.dnn}. It measures the expected error rate per class instead of per dataset. By comparing this measurement with the average error (per dataset), it may suggest different performance on binary data and data with more than 2 classes. For example, compared to WEASEL, mtSAX-SEQL has comparable MCPE but much higher average error (per dataset). This would suggest WEASEL performs better on datasets that have more than two classes.
With regard to average error, Table~\ref{table:sota_stats} suggests FCN is the top performer, but it is followed closely by other methods (mtSS-SEQL+LR, ResNet, WEASEL, COTE).

\begin{table}[htb]
    \sisetup{round-mode=places, round-precision=3}
    \centering
    \caption{Performance of the compared classifiers on the UCR Archive by Mean Per Class Error, Average Classification Error, Number of Wins and Average Rank. As in the critical diagram, the methods are ordered by average error rank.}
   	\label{table:sota_stats}
    \begin{filecontents*}{csv/tsc_stats.csv}
methods,mpce,avgerror,win,avgrank
mtSS-SEQL+LR,0.0424,0.163,16, 6.18
FCN,0.0392,0.156,24, 6.36
COTE,0.0449,0.162,13, 6.69
ResNet,0.0416,0.162,18, 7.45
WEASEL,0.0450,0.167,17, 7.46
mtSFA-SEQL+LR,0.0464,0.174, 8, 7.99
mtSS-SEQL,0.0465,0.183, 9, 8.81
ST,0.0472,0.178, 8, 9.08
mtSAX-SEQL+LR,0.0454,0.177,10, 9.21
BOSS,0.0491,0.186,11, 9.99
mtSAX-SEQL,0.0478,0.192, 9,10.69
mtSFA-SEQL,0.0518,0.197, 5,10.83
EE\_PROP,0.0602,0.207, 8,11.39
LearningShapelet,0.0580,0.213, 7,11.53
TSBF,0.0595,0.222, 6,12.18
MLP,0.0682,0.248, 4,13.22
DTW\_CV,0.0687,0.240, 2,14.44
SFA-SEQL,0.0620,0.259, 4,15.00
SAX-SEQL,0.0657,0.271, 3,15.75
DTW,0.0740,0.262, 5,15.75
\end{filecontents*}
\csvreader[
               table head=\toprule {Classifier} & {MPCE} & {Avg. Error} & {Wins} & {Avg. Rank}\\ \midrule,
               tabular={lSSlS},
               head to column names,
               late after last line=\\\bottomrule]
               {csv/tsc_stats.csv}{}%
     {\methods & \mpce & \avgerror & \win & \avgrank}%
\end{table}

An observable trend for our SEQL-based classifiers is that the more symbolic representations are used, the more accurate the resulting classifier is. 
From right to left in the critical diagram (Figure~\ref{fig:sota1}), we start with a single representation (SAX/SFA), then add more representations of the same type (mtSAX/mtSFA), and finally combine representations of different types (mtSS). 
By combining representations from multiple-resolutions and multiple-domains to create features, we allow the classifier to select only those representations and features that represent the data well, 
and we do not have to decide in advance what are suitable symbolic parameters for the representations.

We note from Table~\ref{table:sota_stats} that SAX and SFA-based classifiers have similar accuracy, with SFA-based once slightly more competitive, depending on how the combination of symbolic representation and SEQL algorithm is done (i.e., using SEQL for ensemble versus feature selection).
The SFA-based algorithm family in the diagram includes WEASEL, BOSS and our SEQL-based classifiers (SFA-SEQL, mtSFA-SEQL and mtSFA-SEQL+LR). Our representative mtSFA-SEQL+LR overtakes BOSS and comes close behind WEASEL. It is worth noting that the output of the SFA transformation used in our experiments is more similar to the transformation used in BOSS, rather than in WEASEL. 
For WEASEL, the authors applied new transformation techniques including non-overlapping windows, bi-gram sequences and full ranges of sliding window size. 
The experiments highlight the potential advantage of our SEQL-based algorithms in exploring the all-subsequence SFA word space, an ability which the BOSS classifier lacks.

On the leftmost of the diagram is our best algorithm (mtSS-SEQL+LR) which utilizes both SAX and SFA representations for the classification task.
 Our method manages to top the average ranking in the pool of state-of-the-art algorithms. 
 With these results, we show that even with a simple linear algorithm, it is still possible to achieve accuracy results comparable to those of very powerful and very complex models, 
 such as large ensembles (COTE) and deep learning approaches (FCN). 
 
\begin{figure}[!ht]
\centering
\includegraphics[width=1\textwidth]{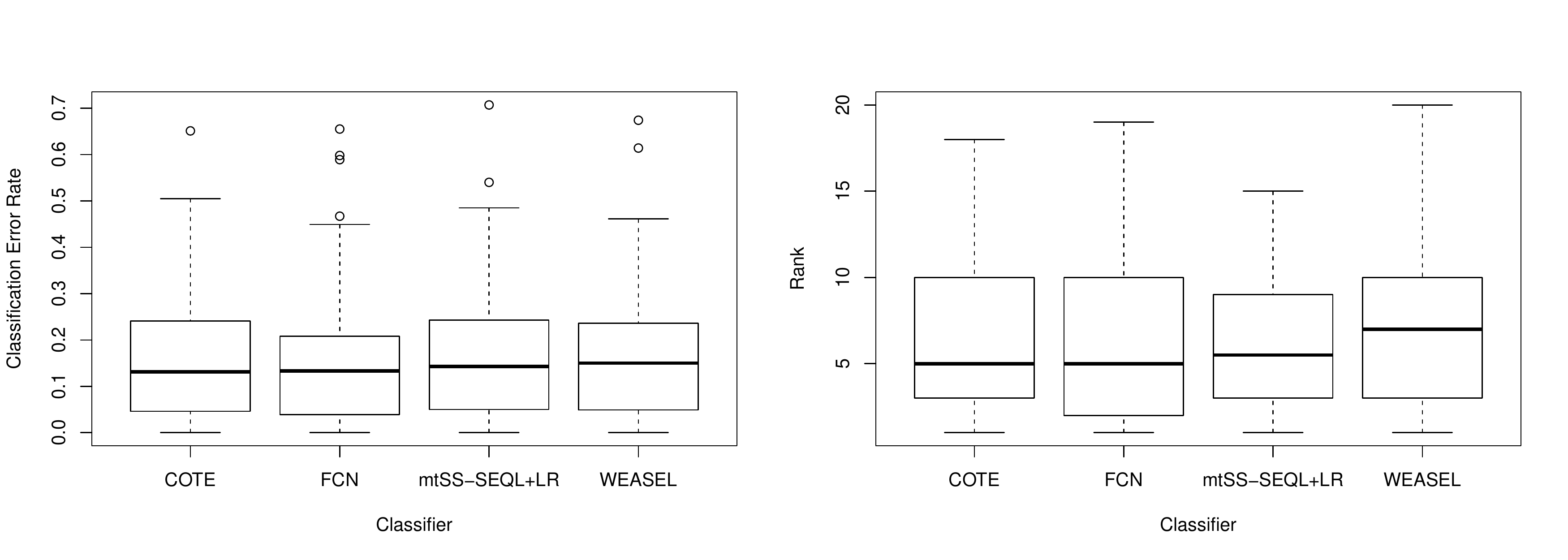}
\caption{Boxplot of classification error and error rank for top-4 methods across all 85 UCR datasets.}
\label{fig:boxplot-error-rank}
\end{figure}

 Since the ranking with regard to average error rank and average absolute error are different, in Figure \ref{fig:boxplot-error-rank} we show box plots of classification error and error rank for the 4 most competitive methods, across 
 the 85 UCR datasets. We note that all 4 methods behave similarly, and all of them have a few datasets where the error is higher than the average behaviour. We discuss this behaviour in the next subsections, in particular we show that 
 different methods are more competitive on different problem types, e.g., FCN does well on images, while mtSS-SEQL+LR does well on motion data.

\subsection{Impact of Multiple Representations}

From the previous experiments, we observe that the strength of our models is derived from multiple symbolic representations of time series. 
However, the multiple representations also bring up the cost of transformation and training. There is also the question of overfitting due to the size of the models. 
Therefore, we study this trade-off by adjusting the number of representations. This can be done by changing the default step size in Algorithm~\ref{alg:ensemble_train} (Line 3) or Algorithm~\ref{alg:seql_as_fs} (Line 4).

\newenvironment{customlegend}[1][]{%
    \begingroup
    \csname pgfplots@init@cleared@structures\endcsname
    \pgfplotsset{#1}%
}{%
    \csname pgfplots@createlegend\endcsname
    \endgroup
}%

\def\addlegendimage{\csname pgfplots@addlegendimage\endcsname}

\begin{figure}[!ht]
\centering
\begin{tikzpicture}
        \begin{customlegend}[legend columns=3,legend style={align=left,draw=none,column sep=2ex},legend entries={mtSAX-SEQL+LR ,mtSFA-SEQL+LR ,Default Configuration,Method D}]
        \addlegendimage{color=blue,mark=*,line legend}
        \addlegendimage{color=red,mark=square*,solid}   
        \addlegendimage{mark=o,mark size = 6pt,mark options={color=black},color=white}
        \end{customlegend}
     \end{tikzpicture}
\begin{tikzpicture}
\begin{axis}[
	name=plot1,
	xlabel={Number of representations},
	ylabel={Classification Error Rate},
	ytick={0.18 ,  0.19}
	]
\addplot table [x=reps, y=acc, col sep=comma] {csv/multireps.csv};
\addplot table [x=reps, y=acc, col sep=comma] {csv/multirep_sfa.csv};
\addplot [mark=o,mark size=6pt,color=red] coordinates {(17.91860465116279,0.173994658)};
\addplot [mark=o,mark size=6pt,color=blue] coordinates {(17.91860465116279,0.1775524497674419)};

\end{axis}
\begin{axis}[
	name=plot2,
	at=(plot1.right of south east), anchor=left of south west,
	xlabel={Number of representations},
	ylabel={Training time in seconds}
	]
\addplot table [x=reps, y=runtime, col sep=comma] {csv/multireps.csv};
\addplot table [x=reps, y=runtime, col sep=comma] {csv/multirep_sfa.csv};
\addplot [mark=o,mark size=6pt,color=red] coordinates {(17.91860465116279,82.7118721163)};
\addplot [mark=o,mark size=6pt,color=blue] coordinates {(17.91860465116279,188.0330947093)};
\end{axis}

\end{tikzpicture}
\caption{ Impact of multiple symbolic representations on TSC: trade-off between accuracy and run time when increasing the number of input symbolic representations. 
All measurements are averaged across the UCR Time Series Archive.}
\label{fig:mtrep}
\end{figure}
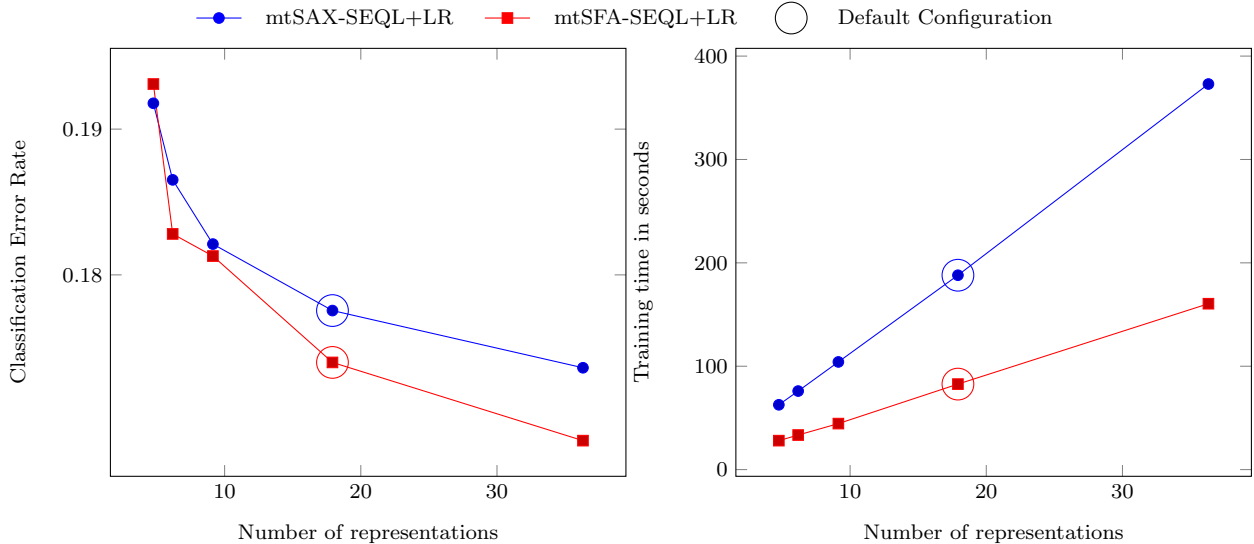

The impact of multiple representations (e.g., muti-resolution for SAX) can be observed in Figure~\ref{fig:mtrep}. In particular, we run the same algorithms but with 5 different step sizes (where $L$ is the time series length, and the step size influences the number of sliding windows): $4\sqrt{L}$, $3\sqrt{L}$, $2\sqrt{L}$, $\sqrt{L}$ (the default configuration) and $\frac{\sqrt{L}}{2}$. Greater step size means fewer representations and vice versa. As expected, adding more representations can improve accuracy, with the extra cost of computation. In fact, the experiment shows that our methods can even achieve better error rate than reported in the previous section. However, we keep the default configuration at $\sqrt{L}$  as suggested in \cite{schafer2015boss} (the BOSS paper) for fair comparison. Moreover, the curves also show that this effect gradually fades out when the representations keep increasing.

\subsection{Accuracy by Data Type}

Following the discussion in \cite{bagnall2016great}, we also measured the performance of our proposed algorithms for each type of data in the UCR Archive. 
In summary, there are 7 types of data: Image Outline (29 datasets), Sensor Readings (16 datasets), Motion Capture (14 datasets), Spectrographs (7 datasets), Electric Devices (6 datasets), ECG measurements (7 datasets) and Simulated (6 datasets).

Table \ref{table:rank_by_problem_type} and \ref{table:error_by_problem_type} report the average error and average error rank of each method and each type of data. Our SEQL-based methods seem to do well on Motion, Spectrograph and Devices data. SFA-based classifiers appear to not be suitable for time series data from Electric Devices. Deep Learning methods (FCN and ResNet) work well across many types of time series with the exception of Motion. COTE has the best performance on Motion and Sensor data, but suffers on Spectrographs. In addition, we can also see the difference between SAX and SFA from the table, by comparing mtSAX-SEQL+LR and mtSFA-SEQL+LR (same classifier but with different input representation). It seems SFA has an advantage in Motion and Image categories while SAX has an advantage in Electric Devices category. Although these results are interesting, we also agree with \cite{bagnall2016great} that the sample sizes are too small to make any definite conclusions.

\begin{table}[ht]
    \centering
    \caption{Average Error Rank by problem type for all compared classifiers. Best method for each type highlighted in bold.}
   	\label{table:rank_by_problem_type}
    \begin{filecontents*}{csv/rank_by_problem_type.csv}
methods,IMAGE,SPECTRO,SENSOR,SIMULATED,ECG,DEVICE,MOTION
mtSS-SEQL+LR, 6.74, \textbf{6.64}, 6.84, 6.17, 5.86, 5.67, 4.43
mtSFA-SEQL+LR, 7.93, 8.71, 8.19, 9.00, 7.86, 9.17, 6.64
mtSS-SEQL, 8.98, 8.00, 9.44,10.17,10.00, 8.92, 6.93
mtSAX-SEQL+LR,10.22, 8.50, 8.66, 8.17, 9.21, \textbf{4.33},10.61
mtSAX-SEQL,10.72, 8.07,11.53,10.83,11.50, 7.83,11.75
mtSFA-SEQL,10.36,10.07,11.38,12.75,10.86,12.08,10.18
SFA-SEQL,14.38,10.29,14.16,18.33,16.93,15.33,17.07
SAX-SEQL,15.43,12.43,15.53,16.67,17.93,12.08,18.39
MLP,11.14,10.07,14.06,14.67, 9.64,19.67,16.54
FCN, \textbf{5.81}, 7.36, 6.25, 7.42, 5.21, 4.50, 8.04
ResNet, 7.57, 7.07, 7.09, \textbf{4.17}, 9.07, 4.83, 9.50
WEASEL, 6.78, 8.64, 7.72, 7.33, 6.36,10.25, 7.39
DTW\_CV,14.57,13.36,14.91,13.33,14.00,16.00,14.18
DTW,15.78,15.57,15.25,11.25,18.93,15.75,16.71
BOSS, 8.47,10.29,12.31,10.83, 9.79, 9.75,10.21
LearningShapelet,12.29,16.21, 8.91,10.75, 9.86,16.83, 9.50
TSBF,11.53,12.57,13.41,10.50,15.14,11.83,11.32
ST,10.97,10.71, 8.00, 9.92, 8.00, 5.00, 7.54
EE\_PROP,11.98,14.29,11.38,10.17, 9.36,13.50, 9.39
COTE, 8.34,11.14, \textbf{5.00}, 7.58, \textbf{4.50}, 6.67, \textbf{3.68}
\end{filecontents*}
\csvreader[
               table head=\toprule {methods} & {IMG} & {SPECTR} & {SENSOR} & {SIMUL} & {ECG} & {DEV} & {MOTION}\\ \midrule,
               tabular={llllllll},
               head to column names,
               late after last line=\\\bottomrule]
               {csv/rank_by_problem_type.csv}{}
     {\methods & \IMAGE & \SPECTRO & \SENSOR & \SIMULATED & \ECG & \DEVICE & \MOTION}%
\end{table}

\begin{table}[ht]
    \centering
    \caption{Average Error by problem type for all compared classifiers. Best method for each type highlighted in bold.}
   	\label{table:error_by_problem_type}
    \begin{filecontents*}{csv/error_by_problem_type.csv}
methods,IMAGE,SPECTRO,SENSOR,SIMULATED,ECG,DEVICE,MOTION
mtSS-SEQL+LR,0.166,0.123,0.156,0.0617,0.0646,0.301,0.218
mtSFA-SEQL+LR,0.173,0.141,0.167,0.0708,0.0737,0.325,0.228
mtSS-SEQL,0.193,0.132,0.169,0.0880,0.0850,0.323,0.230
mtSAX-SEQL+LR,0.186,0.136,0.163,0.0707,0.0721,0.289,0.242
mtSAX-SEQL,0.205,0.136,0.176,0.0880,0.0973,0.317,0.251
mtSFA-SEQL,0.201,0.144,0.183,0.1198,0.0946,0.350,0.247
SFA-SEQL,0.255,0.162,0.213,0.1950,0.1517,0.396,0.391
SAX-SEQL,0.273,0.188,0.240,0.1858,0.1593,0.352,0.404
MLP,0.211,0.194,0.233,0.1688,0.0850,0.539,0.363
FCN,\textbf{0.154},\textbf{0.119},0.142,0.0705,0.0636,0.255,0.236
ResNet,0.170,\textbf{0.119},\textbf{0.140},\textbf{0.0332},0.0820,\textbf{0.254},0.246
WEASEL,0.169,0.128,0.170,0.0527,0.0487,0.342,0.210
DTW\_CV,0.250,0.197,0.222,0.1268,0.1299,0.406,0.294
DTW,0.265,0.231,0.248,0.1297,0.2146,0.400,0.311
BOSS,0.185,0.150,0.195,0.0758,0.0814,0.332,0.234
LearningShapelet,0.229,0.284,0.157,0.0768,0.0834,0.420,0.241
TSBF,0.229,0.191,0.209,0.0900,0.1359,0.393,0.264
ST,0.203,0.135,0.159,0.0710,0.0617,0.294,0.221
EE\_PROP,0.209,0.211,0.196,0.0993,0.0943,0.373,0.247
COTE,0.173,0.153,\textbf{0.140},0.0615,\textbf{0.0433},0.305,\textbf{0.209}
\end{filecontents*}
\csvreader[
               table head=\toprule {methods} & {IMG} & {SPECTR} & {SENSOR} & {SIMUL} & {ECG} & {DEV} & {MOTION}\\ \midrule,
               tabular={llllllll},
               head to column names,
               late after last line=\\\bottomrule]
               {csv/error_by_problem_type.csv}{}
     {\methods & \IMAGE & \SPECTRO & \SENSOR & \SIMULATED & \ECG & \DEVICE & \MOTION}%
\end{table}

\subsection{Running Time}

Table \ref{table:seql_runtime} reports the average running time (in seconds) for each step in our experiments. As it was discussed previously, the independent training of each representation makes it possible to parallelise our algorithm. However, our current implementation is limited to sequential programming. Therefore, beside the total time for training (TotalLearn) and testing (TotalTest) we also report the longest time for training (MaxLearn) and testing (MaxTest) of a single representation per dataset, as the theoretical optimal running time for a parallel implementation. In summary, we show the runtime for the following steps in our algorithms:

\begin{itemize}
\item Transform: Average running time to transform raw data to symbolic representation. Note that we used the authors' implementation \cite{schaefer:dmkd16} for SFA transformation.
\item TotalLearn: Average training time in the case of multiple representations.
\item MaxLearn: Maximum average training time in the case of single representation.
\item TotalTest: Average testing time in the case of multiple representations.
\item MaxLearn: Maximum average testing time in the case of single representation.
\item LogReg: Average training and testing time with logistic regression.
\end{itemize}

\begin{table}[htb]
    \sisetup{round-mode=places, round-precision=3}
    \centering
    \caption{Average running time (seconds) of SEQL-based time series classifiers across all UCR datasets.}
   	\label{table:seql_runtime}
    \begin{filecontents*}{csv/avg_runtime.csv}
methods,transform,totallearn,maxlearn,totaltest,maxtest,vectorspace,logreg
SAX-based,78.77543023255814,188.0330947093,5.2584087674,90.6457399186,3.5902067093,92.8722432674,4.11706976744186
SFA-based,3.7186511627906973,82.7118721163,2.718421093,24.4609997907,1.0373057674,23.5192701512,4.084813953488372
\end{filecontents*}
\csvreader[
               table head=\toprule {methods} & {Transform} & {TotalLearn} & {MaxLearn} & {TotalTest} & {MaxTest}  & {LogReg} \\ \midrule,
               tabular={lSSSSSS},
               head to column names,
               late after last line=\\\bottomrule]
               {csv/avg_runtime.csv}{}%
     {\methods & \transform & \totallearn & \maxlearn & \totaltest & \maxtest & \logreg}%
\end{table}

We attempted to reproduce the experiments by other studies including SAX-VSM, BOSS, WEASEL, FCN and ResNet in order to have a fair comparison in terms of efficiency. However, except BOSS, none of these methods managed to complete the experiment in reasonable time with our machine (a desktop personal computer). The current implementation of BOSS is the most efficient implementation among the top classifiers to the best of our knowledge. On the other hand, WEASEL demands a lot of memory which the system failed to provide in the case of bigger datasets. We suspect its super rich feature space and the absence of effective pruning techniques are the causes for this huge demand of computing resources.

\section{Interpretability}
\label{sec:interpretability}
As described in Section~\ref{sec:method}, the output of SEQL is a linear model (a weighted list of selected features) which makes interpretation possible. 
In this section, the focus is on the interpretation in the context of TSC, i.e., how to identify the segments that are important for the classification decision. 
Since we visualise the data in the time domain, we only discuss SAX-SEQL-based models here. Regarding SFA representations, there have so far been no results that report whether SFA sequences are interpretable. 
We visualize SAX sequences by mapping each SAX word back to its corresponding segment in the original time series. 
Technically, the same mapping can be done with SFA sequences, but visualising the impact of such features in the time domain is not suitable.

\subsection{Feature Importance}

In our previous study~\cite{fvseql:7930038}, we evaluated the importance of the features selected in the final model by studying the coefficients learned by SEQL. 
Basically, the coefficient of a feature can imply which class the feature represents (based on the sign) and how decisive the feature is in the classification decision (based on the absolute value). 
For multiple representations, a feature is defined not only by the sequence but also by its SAX parameters. 
Table~\ref{table:gun_point_top10} shows some of the features selected by the mtSAX-SEQL+LR algorithm. 


\begin{table}[htb]
    \sisetup{round-mode=places, round-precision=5}
    \centering
    \caption{Top 10 features selected by mtSAX-SEQL+LR from the Gun\_Point time series dataset.}
   	\label{table:gun_point_top10}
    \begin{filecontents*}{csv/gun_point_top10.csv}
l,w,ad,coef,seq
42,16,4,0.06584011,cbaab
53,16,4,0.0624718,db
53,16,4,0.062228679,ddddb
42,16,4,0.06199756,da
31,16,4,0.059719777,bbbbbbbbbbcdddd
53,16,4,-0.053721338,aaaaaabbbb
20,16,4,-0.054386522,bbbbaaaaaa
53,16,4,-0.054575478,bbbcddddd
53,16,4,-0.05574525,bbbbbbbaaa
53,16,4,-0.060501812,bbbbbbaaa
\end{filecontents*}
\csvreader[
               table head=\toprule {$l$} & {$w$} & {$\alpha$} & {Coefficients} & {Subsequences}\\ \midrule,
               tabular={lllSl},
               head to column names,
               late after last line=\\\bottomrule]
               {csv/gun_point_top10.csv}{}%
     {\l & \w & \ad & \coef & \seq}%
\end{table}

\subsection{Visualizing SAX Features}

\begin{figure}[h]
\begin{center}
\includegraphics[width=0.8\textwidth]{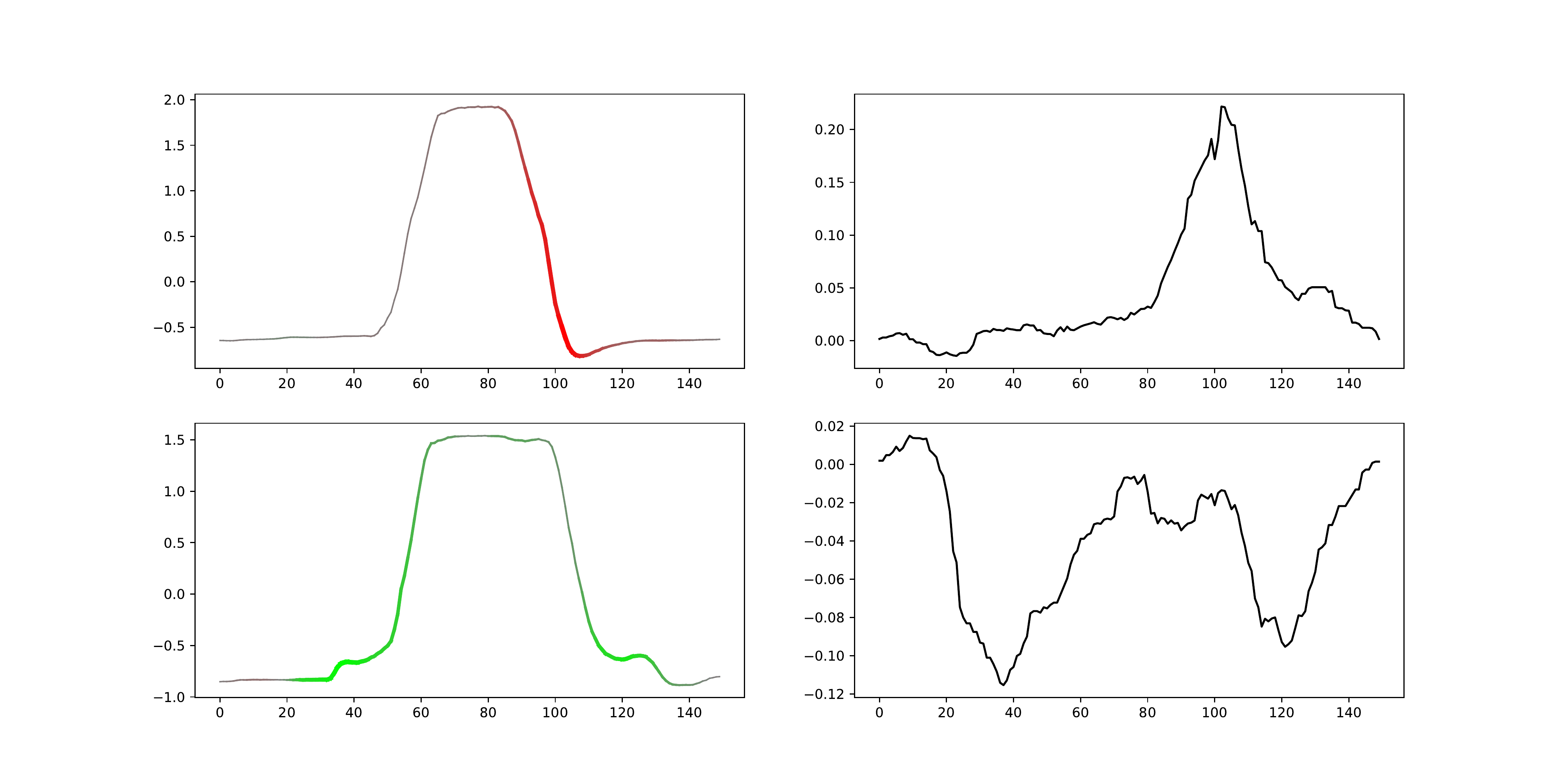}
\caption{Example from the Gun\_Point dataset. The top and bottom left plots illustrate the time series with highlighted regions (top is Point). The top and bottom right plots measure the classification weight of each point in the time series.}
\label{figure:gunpoint_example}
\end{center}
\end{figure}

Two examples of time series from the Gun\_Point dataset can be found in Figure~\ref{figure:gunpoint_example}. The time series records the motion of the hand when pointing (Point class) or drawing a gun (Gun class). The time series from the Gun class (bottom-left) is characterized by two little perturbations at the beginning and the end of the motions. On the other hand, the time series from the Point class (top-left) is characterized by the small dip that occurs when the motion ends. The highlighted regions were discovered by our SEQL classifier by mapping the matched features to the raw time series.

Top-right and bottom-right are the visualizations of the classification decision. We calculated the importance of each point in the time series based on the coefficients of the features, i.e., if a feature is mapped to a segment of a time series, then each point in that segment receives a proportional value from the coefficient (Algorithm~\ref{alg:mapsax}). Therefore, a peak means the corresponding region in the original time series strongly suggests the positive class and a valley means the corresponding region strongly suggests the negative class. The result is a ``meta time series'' which contains a classification description of the classification model on a time series. 
The peaks and the valleys are also reflected in the highlighted part of the time series plots (top and bottom-left).

\begin{algorithm}[h]
\caption{Mapping symbolic features back to the original time series.}
\begin{algorithmic}[1]
\Function{findSegments}{feature, timeseries}
\State Initialize the meta time series $mtts = zeros(lengthof(timeseries))$ 
\State $sax = SAXtransform(timeseries,feature.l,feature.w,feature.\alpha)$  
\State Find all the locations of $feature.sequence$ in $sax$.
\ForAll{$loc$ in $locations$}
\State $mtts[loc] += feature.coef/locations.size()$
\EndFor
\State \Return $mtts$
\EndFunction
\end{algorithmic}
\label{alg:mapsax}
\end{algorithm}

\begin{figure}[htb]
\begin{center}
\includegraphics[width=0.7\textwidth]{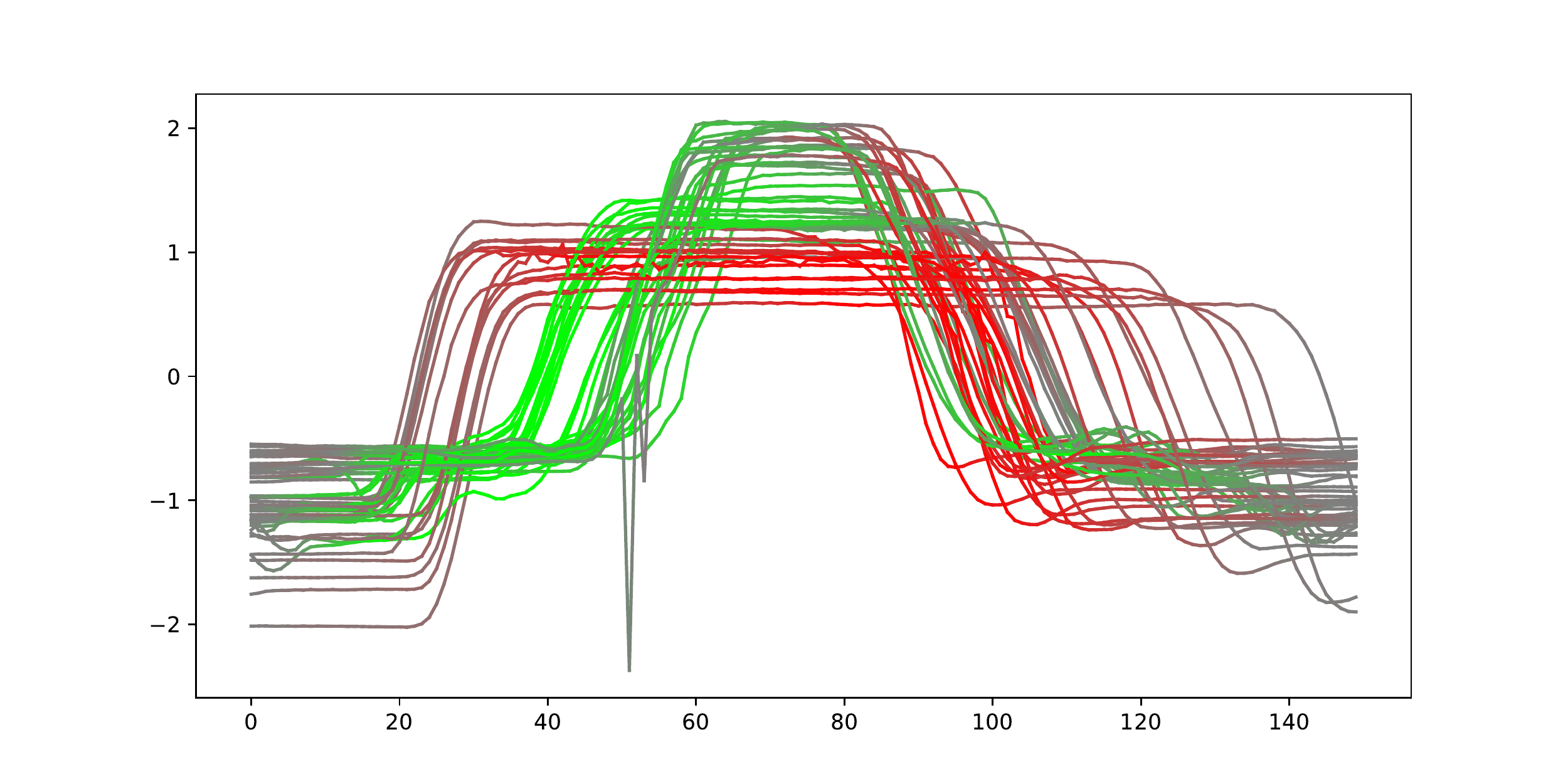}
\caption{Visualising important segments for the classification decision across all Gun\_Point time series (best seen in color).}
\label{fig:gun_point_all}
\end{center}
\end{figure}

In Figure~\ref{fig:gun_point_all}, we plot all the time series in the dataset together, to find the overall trend. From that figure, it seems that the classification model tends to focus more on the regions describing the gun drawing motion, since the green highlights dominate this area of the Gun-class time series. On the other hand, the model found the final dips of the Point-class time series to be the most reliable feature, as indicated by the red highlights.



\begin{figure}[h]
\begin{center}
\includegraphics[width=0.8\textwidth]{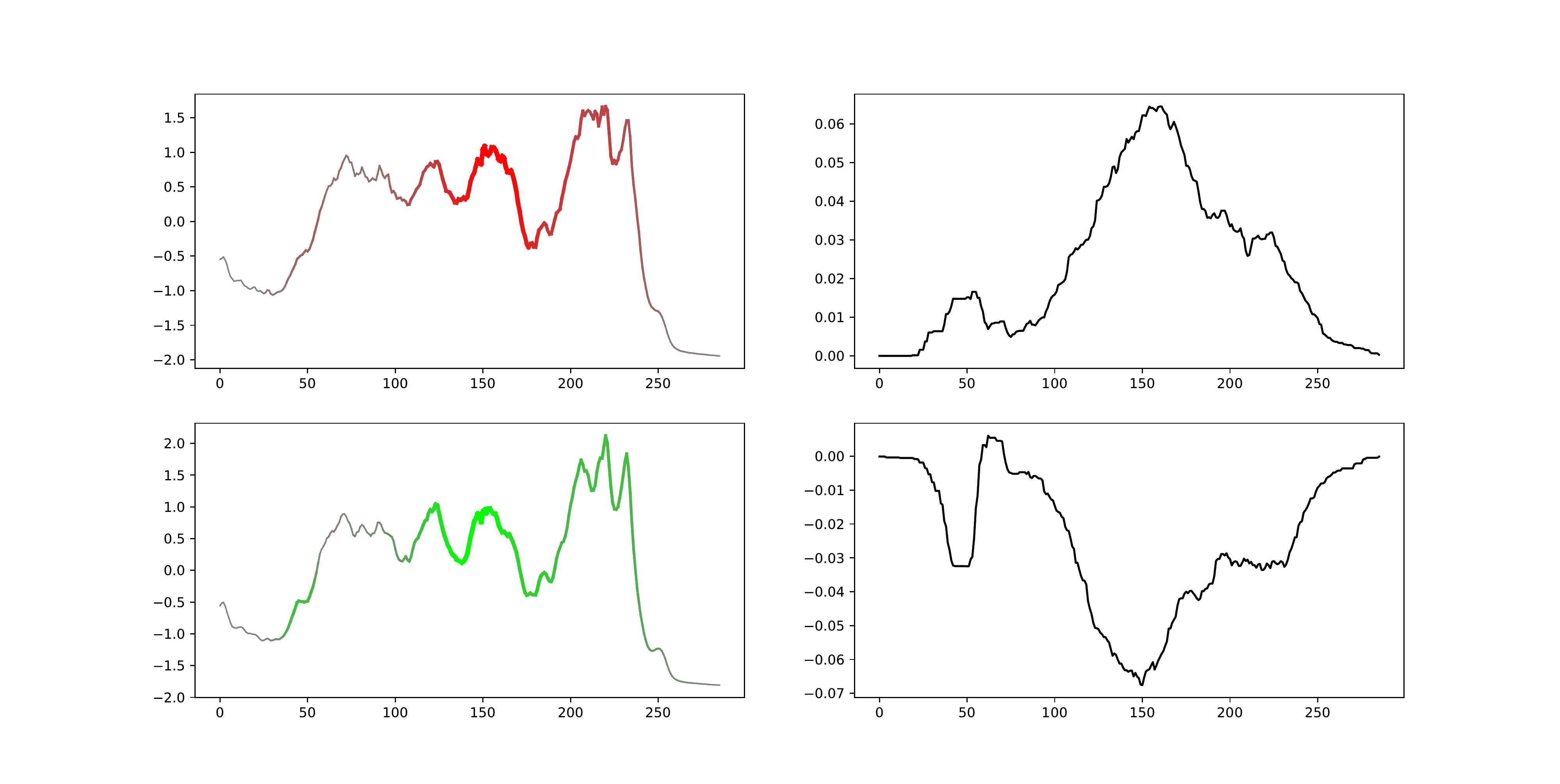}
\caption{Example from the Coffee dataset. The top and bottom left plots illustrate the time series with highlighted regions (top is Arabica). The top and bottom right plots measure the classification weight of each point in the time series. }
\label{figure:coffee_example}
\end{center}
\end{figure}

Figure~\ref{figure:coffee_example} presents another two examples from the Coffee dataset: one from the Arabica class (top-left) and one from the Robusta class (bottom-left). By observing the highlights and the ``meta time series'', it is fairly clear how the classifier made the decision. In fact, the highlighted regions correspond to the caffeine and chlorogenic acid components of the coffee blends~\cite{coffee}.

\section{Case Study: TSC Interpretability on the FLUXNET Benchmark}

In order to further explore the importance of the interpretability aspect in TSC in a real world set-up, we work with an ecology benchmark and compare some of the state-of-the-art time series classification approaches.
 The studied methods are: 1NN-DTW, BOSS, WEASEL, mtSAX-SEQL+LR and CMAGEP \cite{Ilie2017a} (a symbolic regression method that returns a learned formula that best explains the data). 
The comparison is done based on class label prediction accuracy, time efficiency and interpretability. 

The real world problem we set to solve is a binary time series classification problem from the biogeochemistry field, where the purpose is to accurately determine the climate type in which time series of 
ecosystem carbon fluxes are measured. This is a binary classification task, in that only two types of climate are considered, i.e., temperate and non-temperate climate, 
and all the measurement sites should be mapped to either the former or the latter. 

A very interesting aspect of this problem is that interpreting the classification decision is desired, as the underlying scientific question we are trying to answer here,
 is whether ecosystem carbon fluxes have different underlying processes depending on the climate type of the ecosystem.

%
%

\subsection{Data Description and Preparation}

Our benchmark study is based on measurements from the La Thuile FLUXNET `fair use' database\footnote{\url{http://www.fluxdata.org}}. FLUXNET \cite{Baldocchi2008} is a global spatially distributed network of measurement sites that continuously monitor and record environmental biotic and abiotic variables. Some examples of such variables are temperature, precipitations, carbon, water and energy fluxes. In the current study we focused on the daily aggregate of terrestrial ecosystem carbon outgoing flux, also called terrestrial ecosystem respiration ($R_{eco}$)  \cite{Reichstein2005,Baldocchi2003}. 

From the total of 152 available measurements sites in this database, we selected 96 sites based on completeness of records. 
Although in all studied sites the available time series were spreading over more than 1 year, here we focused on the year 2005 as it contained the more complete $R_{eco}$ time series over-all sites. 
Each site contains a detailed description of the local environmental conditions, among which a climate type label, allowing for a straightforward training and validation against the truth.
For the classification benchmark, we shuffled the sites and did a stratified split by class types into training and test sets. 
The training set contained 65 $R_{eco}$ time series (31 in temperate climate and 34 in non-temperate climate), 
while 31 $R_{eco}$ time series were kept for test data (14 temperate climate sites and 17 non-temperate climate sites). 

%

\subsection{Method and Experimental Set-up}

We ran 1NN-DTW, WEASEL, BOSS (Ensemble and VS) with default settings and mtSAX-SEQL+LR  (with settings as in Table \ref{table:params}) 
to train time series classification models on the $R_{eco}$ measurements from the 65 sites in the training sample, 
and the resulting test classification errors and individual run times were reported in Table \ref{tab:resultsBenchRespitation}.

We have also fitted CMAGEP models \cite{Ilie2017a} for learning formulas that were further used in classification. CMAGEP is a recently proposed hybrid algorithm 
that combines a genetic programming approach (GEP \cite{Koza1994,Ferreira2004}) for automatic function discovery in symbolic regressions, 
with an evolutionary strategy approach (CMA-ES \cite{Hansen2003}) for local optimization. 
CMAGEP was shown to slightly surpass the original GEP in prediction performance, and to return much shorter solutions (i.e., formulas) than the original GEP, allowing for ease in interpretation.
Here, the CMAGEP mathematical model formulations were built starting from $R_{eco}$ and its time lags for the sites in each climate type as follows:
\begin{enumerate}
	\item Stack all same-labelled training site time series and generate daily lags up to 4 days behind;
	\item Use lagged $R_{eco}$ as input dependent features and the current $R_{eco}$ values as target;
	\item Randomly generate initial set of candidate solutions based on input and CMAGEP parameter settings;
	\item Compute fitness of solutions based on CMAGEP parameter settings;
	\item CMA-ES locally optimize set of solutions as given in CMAGEP parameter settings;
	\item Apply genetic variation operators to select candidate solutions with rates given in Table \ref{tab:setCMAGEP} to generate new population of solutions;
	\item Repeat 4-6 or return final solution when stop criterion is reached (best fitness, maximum run time, maximum unimproved generation count, etc.)
\end{enumerate}

Once the two model formulas are available (one for the positive training set and one for the negative one), the decision for classifying a test site time series to one of the two climate types, is based on the 
highest CMAGEP model fit computed using R square \cite{Ilie2017a}. 

\begin{table}[h]
	\small
	\centering
	\caption{Parameter settings for each individual CMAGEP run.}
	\label{tab:setCMAGEP}
	\begin{tabular}{ll}
		\hline
		Parameter & Value\\ \hline
		No. of runs &20\\
		Number of chromosomes &200\\
		Number of genes & 3\\
		Head length &5\\
		Functions & $+,-,/,*,x^{y},\sqrt,\ln,\exp$\\
		Terminals & $R_{eco}(t-1),R_{eco}(t-2),R_{eco}(t-3),R_{eco}(t-4)$ \\
		Link function & +\\
		Max run time &3600 seconds\\
		Fitness function & CEM \cite{Ilie2017}\\
		Selection method for replication & tournament\cite{Coello2002}\\
		Mutation probability& 0.5\\
		IS and RIS transpositions probabilities &0.05 \\
		Two-point recombination probability & 0.1\\
		Inversion probability & 0.05\\
		One point recombination probability & 0.3\\ \hline
		Start generation for CMA-ES optimization &0\\ \hline
		Maximum iterations for CMA-ES& 10\\ \hline
		Chromosomes to optimize in a generation& 10\\ \hline
	\end{tabular}
\end{table}

\subsection{Results}

After training and prediction 
we found that the lowest test classification error was obtained using 1NN-DTW, followed by CMAGEP and mtSAX-SEQL+LR, with WEASEL and BOSS in a close range of each other. In terms of run time, the five time series classification approaches are in a similar range of each other, with BOSS VS the fastest and 1NN-DTW the slowest. The symbolic regression approach, CMAGEP, needs much longer time to reach a solution as compared to the other methods. 
Once the CMAGEP models are generated based on training data, the time needed to produce predictions is equal to the time it takes to evaluate the mathematical formulas, so a total train+prediction time might be more appropriate for CMAGEP.

The CMAGEP learnt mathematical formulas for the models describing the processes driving the $R_{eco}$ are given in Equation \ref{eq:TempResp}, for the sites in the temperate climate and in Equation \ref{eq:NTempResp} for the sites in the non-temperate climate, respectively ($t$ is the time component of the studied $R_{eco}$ signal in days). 

\begin{equation}
\label{eq:TempResp}
R_{ecoT}(t)=1.2R_{ecoT}(t-1)+0.04R_{ecoT}(t-3)-0.7
\end{equation}

\begin{equation}
\label{eq:NTempResp}
R_{ecoNT}(t)=4.9R_{ecoNT}(t-4)+2.2
\end{equation}

\subsection{Interpretability}

\begin{figure}[ht]
	\centering
	\includegraphics[width=1\textwidth]{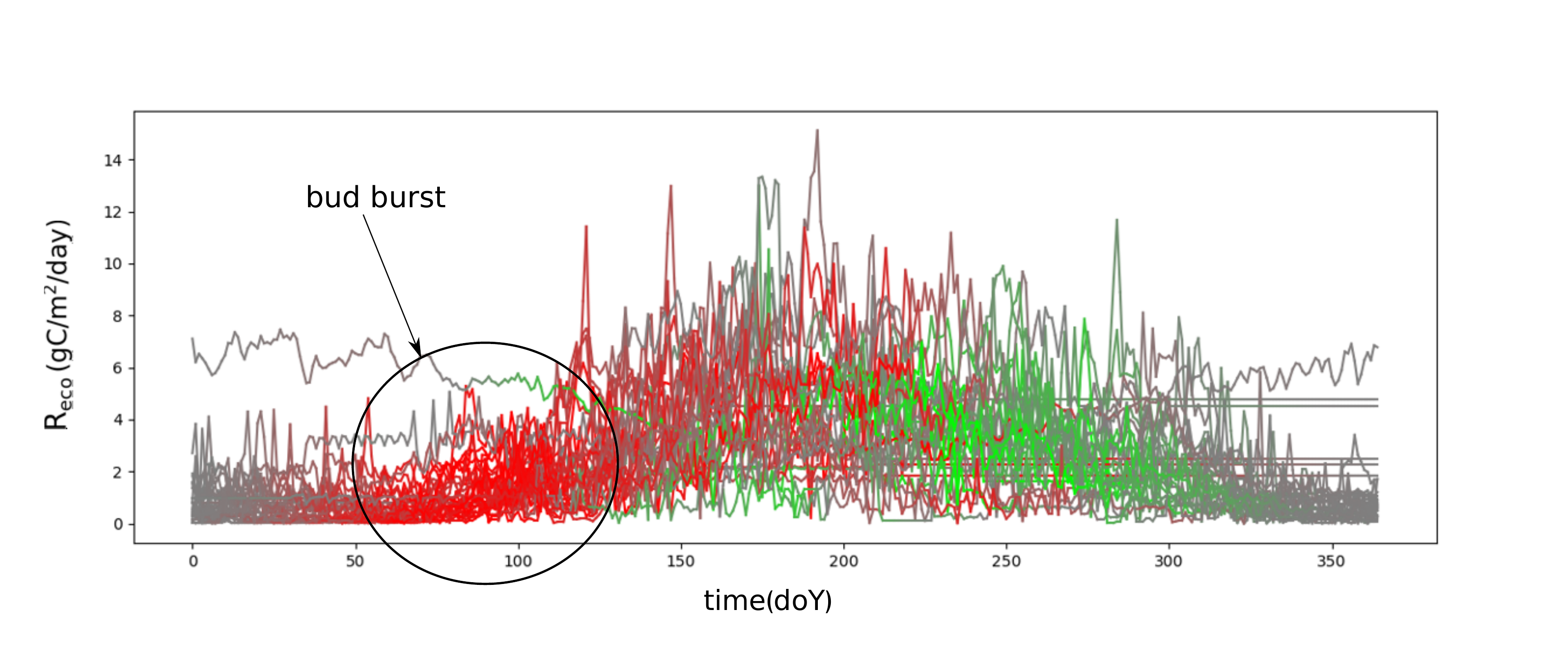}
	\caption{Daily terrestrial ecosystem respiration for the year 2005 in temperate climate sites, with mtSAX-SEQL+LR classification based highlighted regions. 
	The stronger the colour, the higher the weight for the feature pointing to the time series region. 
	Temperate climate class regions are shown in red colour, and non-temperate ones are shown in green.} 
	\label{fig:Temp1YearResp}
\end{figure}

\begin{figure}[ht]
	\centering
	\includegraphics[width=1\textwidth]{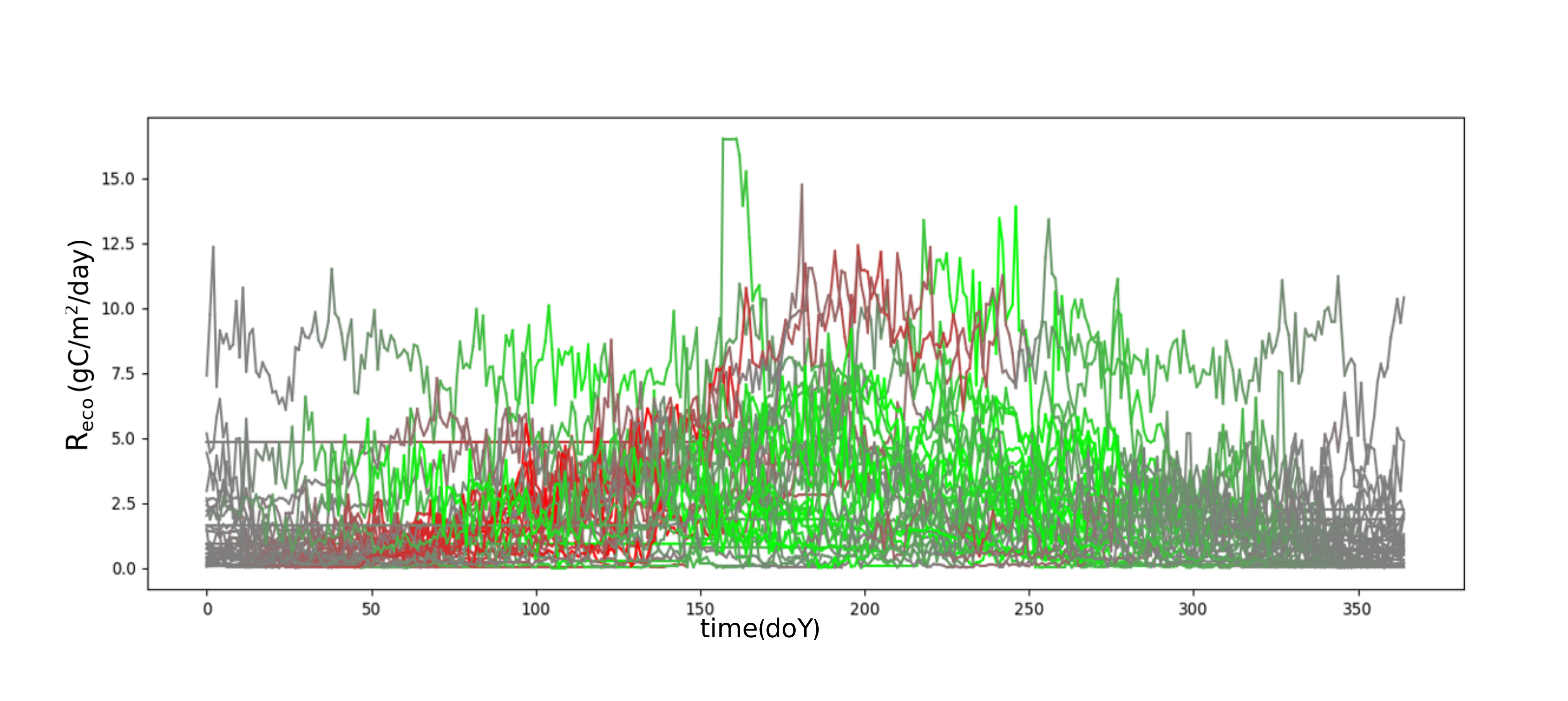}
\caption{Daily terrestrial ecosystem respiration for the year 2005 in non-temperate climate sites, with mtSAX-SEQL+LR classification based highlighted regions. 
The stronger the colour, the higher the weight for the feature pointing to the time series region. Temperate climate class regions are shown in red colour, and non-temperate ones are shown in green.} 
	\label{fig:NTemp1YearResp}
\end{figure}

Among the methods compared, mtSAX-SEQL+LR and CMAGEP can provide meaningful insights into the features and mechanisms that determine the classification decision.
For interpretation, as described in Section \ref{sec:interpretability}, mtSAX-SEQL+LR offers the possibility of explaining the classification decision, 
by allowing to visualise the logistic regression features that map to regions in the time series.
Figures \ref{fig:Temp1YearResp} and \ref{fig:NTemp1YearResp} highlight the most significant regions in the $R_{eco}$ time series for mtSAX-SEQL+LR classification, for all temperate and non-temperate sites.
These figures illustrate the mapping of the features selected by the mtSAX-SEQL+LR models, showing specific regions of the time series that are more significant in revealing the climate type where a certain $R_{eco}$ flux was measured. 

\begin{table}[ht]
	\centering
		\caption{Classification performance for the benchmarked methods on the $R_{eco}$ to climate type mapping problem.}
	\label{tab:resultsBenchRespitation}
	\begin{tabular}{lccl}
		\hline
		Method							& Error 	& Runtime 	& Interpretation\\ \hline
		1NN-DTW	 (\cite{bagnall2016great})		&0.22	&24 seconds	& NO\\
		CMAGEP	(\cite{Ilie2017a}	)			&0.26	& 30 minutes	& YES, explicit model formulation\\
		mtSAX-SEQL+LR (this paper)			&0.32	&11 seconds	&YES, SAX features visualization\\
		WEASEL (\cite{Schafer:2017:weasel})	&0.48	&6 seconds	& NO\\
		BOSS VS	(\cite{schaefer:dmkd16})		&0.52	&1 second	& NO\\				
		BOSS 	(\cite{schafer2015boss})		&0.58	&7 seconds	& NO
	\end{tabular}
\end{table}

Interestingly, the insight we could infer from mtSAX-SEQL+LR models generally coincides with already established knowledge from the ecology field.
Firstly, we could see that the more significant regions of temperate $R_{eco}$ appear to be around the 2-4 month time stamps. In ecology, it is known that the months corresponding to the start of spring is when the bud-burst takes place in the majority of temperate regions, with the bud-burst leading to a strong shift in the plant regime of intake and release of $CO_2$ \cite{Migliavacca2011}. Such a change in respired carbon was also visible in the $R_{eco}$ time series shown in Figure \ref{fig:Temp1YearResp}. 
Secondly, in the non-temperate sites, the most significant regions of the time series for the mtSAX-SEQL+LR classification decisions are shown to be around the 6-9 month time stamps, which coincides with the time of largest peaks followed by a strong decrease in temperature, and implicitly of $R_{eco}$ in the temperate sites during summer and early autumn months. Such strong shifts are however missing in non-temperate climate sites, with $R_{eco}$ showing local variability but no strong seasonal patterns in non-temperate climate sites \cite{Jung2011}  as seen as well in Figure \ref{fig:NTemp1YearResp}. 

CMAGEP on the other hand, being a GP system, gives a different perspective to interpretation and returns human readable formulas. These formulas allow for mathematical and physical interpretation of the studied system processes, possibly leading to discovery of novel knowledge \cite{Ilie2017}. Here, the structures of the CMAGEP models show that there might indeed be different intrinsic mechanisms driving the respiration process in different climate types, such as stronger short term variability in the temperate sites and longer presence of autocorrelations in the non-temperate sites, with $R_{eco}$ being largely dependent on the first lag and less so on the third lag in temperate sites and solely dependent on the fourth lag in the non-temperate sites. 

Based on our interpretation of the $R_{eco}$ responses captured in the mtSAX-SEQL+LR and CMAGEP models, it seems that the studied $R_{eco}$ time series closely follow known temperature and precipitation patterns and seasonalities \cite{Jung2011}, however in order to fully understand how and up to which degree, further investigation is needed that surpasses the illustrative purpose of the current case study.

In the present case study we noticed that depending on the approach we use for inferring knowledge with TSC model building, the processes driving a physical system can be analysed at macro level with mtSAX-SEQL+LR models or it can be zoomed-in with CMAGEP derived models. This leads us to believe that depending on the purpose of the study, mtSAX-SEQL+LR or CMAGEP would be more appropriate, or it could even be that sometimes a combination of the two approaches shown here is the way to go, with mtSAX-SEQL+LR pointing the relevant TSC regions and CMAGEP possibly digging deeper to why the specific regions are significant.

The current results show once more that in order to fully grasp and exploit the importance of interpretabiliy in the context of machine learning modelling, expert knowledge in the studied field is necessary, and that although data alone can help to describe functionalities, if a deeper understanding of the studied systems is sought, data scientists should work side-by-side with domain experts.

With interpretability of TSC showing such a large potential not only for simulating but also for developing a better understanding of real physical systems, 
one of the natural steps that we believe we should take is to try to build approaches that will look at systems responses not only from a univariate perspective, 
but that capture multivariate responses at unique time stamps, because, as seen here, the uni-dimensional case is often insufficient for extracting more than just descriptive information.  
For future research on this benchmark and case-study, we plan to build (multivariate) CMAGEP and mtSAX-SEQL+LR TSC models that learn from other measured fluxes at the FLUXNET sites  (e.g., temperature, humidity, $R_{eco}$).

\section{Conclusion}
\label{sec:conclusion}

The goal of this study is to explore the impact of combining time series symbolic representations and linear classifiers. In this area, it has been commonly perceived that DTW-based methods are hard to beat and ensemble methods (and lately deep learners) often offer the best accuracy. However, the series of SFA-papers \cite{Schafer:2012:SSF:2247596.2247656,schafer2015boss,schaefer:dmkd16,Schafer:2017:weasel} has shown otherwise, in particular the WEASEL method is an effective algorithm that uses a linear classifier in a large-dimensional SFA-words space.
Moreover, our concerns about interpretability and efficiency mean that even the most accurate classifiers can still be less desirable. 
With this paper, we show that linear classifiers can still be strong competitors and symbolic representations are a powerful tool for time series analysis. 

In summary, we study two notable symbolic representations of time series (SAX and SFA) and propose a time series classification framework that can utilize both at different resolutions. 
Due to its pruning ability, our core classifier (SEQL) can navigate the vast symbolic-words space efficiently. We showed that approximation at multiple resolutions is an effective approach, 
instead of trying to find an optimal representation. Furthermore, our classifiers can work with different symbolic representations, thus effectively learn from a multiple domain feature space 
without the need of incorporating various learning algorithms. 
We think that this characteristic has great potential as our classifier can theoretically accommodate other symbolic representations in the future. 
In practice, this flexibility means representations and resolutions can be chosen according to the problem. 

We proposed 8 different SEQL-based algorithms using the SAX and SFA symbolic representations.
To showcase our contributions, we tested our proposals with the full UCR Time Series Archive and demonstrated that they are strongly competitive against the state-of-the-art,
 including against powerful learners such as large ensembles (COTE) or deep learning methods (FCN). 
 While ensemble and deep learners are well-known for their accuracy, they are also notorious for their high demand of computing resources. 
 On the other hand, our SEQL-based methods are more efficient due to the effective combination of symbolic representations and learning algorithm.
The time complexity of our algorithms also enables them to scale well for large datasets. 
The outcome of a SEQL-based TSC algorithm is a linear model, which enables us to interpret the classification decision, a property which is desirable for time series analysis. 
Various ways of interpreting the resulting models were also discussed in this paper, in particular we discussed interpreting the classification decision on well known UCR problems, as well as presented a case-study on an ecology benchmark 
where we can relate the algorithmic decisions to real-world domain knowledge. For the future, we see a lot of potential on extending these methods for multivariate time series classification problems, as well as continuing to work 
on human-friendly ways of explaining the classification decisions.


\section*{Acknowledgment}

This work was funded by Science Foundation Ireland (SFI) under grant number 12/RC/2289.

\bibliographystyle{IEEEtran}
\bibliography{tscseql}
\end{document}